\documentclass{article}
\usepackage{amsmath,amssymb}
\usepackage{booktabs}
\usepackage{multirow}
\usepackage{xspace}
\usepackage{tikz}
\usetikzlibrary{positioning,arrows.meta,fit}
\usepackage{pgfplots}
\pgfplotsset{compat=1.18}
\usepackage[onecolumn]{trustai4s}
\usepackage{hyperref}                % load after trustai4s
\usepackage{xurl}
\newcommand{\sys}{\textsc{Tandem}\xspace}        % paper name (the internal project name is not used in the paper)

\newcommand{\secref}[1]{\S\ref{#1}}
\newif\ifpn \pntrue

\newif\ifupd \updtrue

\newcommand{\zm}[1]{\textcolor{black}{#1}}
\providecommand{\Description}[2][]{}

\labpreprintbrand{}{SMU-FIU TrustAI4S Lab}{}
\mlsystitlerunning{Tandem: Serving Hybrid Linear-Attention GUI Agents on Mobile NPUs}
\makeatletter
\renewcommand{\Notice@String}{Preprint.}
\makeatother
\hypersetup{pdftitle={Tandem: Serving Hybrid Linear-Attention GUI Agents on Mobile NPUs},
  pdfauthor={Zhuomin Chen, Jingchao Ni, Xu Zheng, Yong Chen, Mo Sha, Wei Cheng, Dongsheng Luo}}

\begin{document}
\twocolumn[
\mlsystitle{Tandem: Serving Hybrid Linear-Attention GUI Agents on Mobile NPUs}
% Author order supplied by the authors on 2026-10-16; Xu Zheng and Yong Chen added (Xu Zheng: xzhen019@fiu.edu).
\begin{mlsysauthorlist}
\mlsysauthor{Zhuomin Chen}{fiu}
\mlsysauthor{Jingchao Ni}{uh}
\mlsysauthor{Xu Zheng}{fiu}
\mlsysauthor{Yong Chen}{ind}
\mlsysauthor{Mo Sha}{fiu}
\mlsysauthor{Wei Cheng}{nec}
\mlsysauthor{Dongsheng Luo}{smu}
\end{mlsysauthorlist}
\mlsysaffiliation{fiu}{Florida International University, Miami, FL, USA}
\mlsysaffiliation{uh}{University of Houston, Houston, TX, USA}
\mlsysaffiliation{ind}{Independent Researcher}
\mlsysaffiliation{nec}{NEC Laboratories America, Princeton, NJ, USA}
\mlsysaffiliation{smu}{Singapore Management University, Singapore}
\mlsyscorrespondingauthor{Dongsheng Luo}{luodongsheng01@gmail.com}
\vskip 0.3in
\begin{abstract}
% Abstract, revision 2 (2026-09-29), written to STYLE_MOBISYS.md sec. 3. \pn marks numbers to re-measure on the final
% build (cooler, best configuration of every baseline).
% src: figures/comparison.tex (final build rt25 with all mechanisms, 2026-10-02 evening, screen verifiably on, without
%      log-probabilities): Tandem 1.377 s text / 4.76 s screenshot; llama.cpp NPU without MTP (its fastest) 1.957 /
%      5.66 s; with MTP 2.55 / 7.47 s; GPU and CPU engines 4.98-11.92 s text (3.6-8.7x) and 19.1-67.2 s screenshot
%      (4.0-14.1x). Hybrid vs full attention (tab:model): 43% slower under llama.cpp (1.901 / 1.333 s), 1% under Tandem
%      (1.313 / 1.294 s). Energy: tab:energy (exp_energy_tail.sh, 2026-10-02, four alternating runs each, 10.1 vs 15.1 J
%      per text step on battery, 32.9% less). BASELINE_STACKS.md sec. 1: no stack other than llama.cpp runs the Gated
%      DeltaNet layers on the NPU.
% Percent reductions use 100*(1 - Tandem/baseline): text 29.67%, screenshots 15.94%, rounded to 30% and 16%.
Smartphone GUI agents complete tasks in mobile apps by repeatedly reading the screen and asking a model for the next action. Mobile neural processing units (NPUs) enable local execution, keeping screen contents on the phone, but model latency accumulates with every action. 
Hybrid linear-attention models reduce memory requirements, yet reusing their recurrent state requires checkpoints at the appropriate input positions. Agent requests retain instructions but rewrite recent history, so checkpoint policies designed for chat continuation can miss reusable prefixes. Retaining rollback states for speculative decoding can also move prompt processing off the NPU's matrix unit, adding overhead that outweighs decoding savings on short action outputs. A loaded model can further hold the NPU awake between tasks. 
We present \sys, a runtime for hybrid GUI agents on mobile NPUs. \sys saves recurrent state where consecutive prompts diverge, allowing later requests to reuse previously processed context. It keeps most prefill on the matrix unit and produces rollback states through a short per-token tail. Its verification kernel reads each weight block once for all proposed tokens. Between request bursts, \sys releases NPU power votes without unloading the model. We implement \sys in llama.cpp and evaluate it on recorded agent interactions across two Snapdragon generations. On a Snapdragon 8 Elite, \sys reduces per-step latency by 30\% for text-based screen descriptions and 16\% for screenshots relative to llama.cpp's fastest NPU configuration, reduces energy per text step by 33\%, and lowers standby power with the model loaded from 0.76 to 0.22~W. 
The project is available at \url{https://github.com/flyingdoog/tandem}.

\end{abstract}
]
\printAffiliationsAndNotice{}

% Introduction, revision 2 (2026-09-29), written to STYLE_MOBISYS.md sec. 4 (hook, model, measured costs and their root
% cause, mismatched assumptions of existing systems, key idea, challenges mapped to techniques, results, contributions).
% \pn marks numbers to re-measure on the final build (cooler; best configuration of every baseline).
\section{Introduction}\label{sec:intro}

GUI agents let users operate phone apps through natural-language instructions. A request to change an app setting, for example, may require opening a menu, locating an option and confirming the change. At each step, the agent reads the screen and asks a model to choose a tap, scroll or text input~\cite{wen2024autodroid,wen2025autodroidv2,liu2024autoglm,qin2025uitars}. It then waits for the app to respond before deciding what to do next. 
% Model latency therefore accumulates throughout the task. 
\zm{Model latency accumulates throughout the task, while pauses between requests and tasks also make idle power relevant.}
Running the model on the phone keeps screen contents local and removes a network round trip from each step; mobile neural processing units (NPUs) provide the compute for this execution~\cite{xu2025llmnpu,yin2026shadownpu}.

Hybrid linear-attention models make longer interactions possible within a phone's memory budget. Models such as Qwen3.5-4B replace most attention layers with recurrent layers that summarize earlier input in a fixed-size state~\cite{qwen2026qwen35,yang2025gated}. This slows memory growth as prompts lengthen. 
% Efficient agent execution also requires the runtime to preserve useful states across changing requests and produce the intermediate states needed during action generation. Checkpoint placement and NPU execution determine the cost of these two uses of state.
\zm{Efficient agent execution requires two distinct uses of recurrent state: reusable checkpoints across changing requests and short-lived rollback states for speculative generation within a request. Checkpoint placement and NPU execution determine the cost of maintaining both.}

\begin{figure*}[t]
    \centering    
    \includegraphics[width=0.9\linewidth]{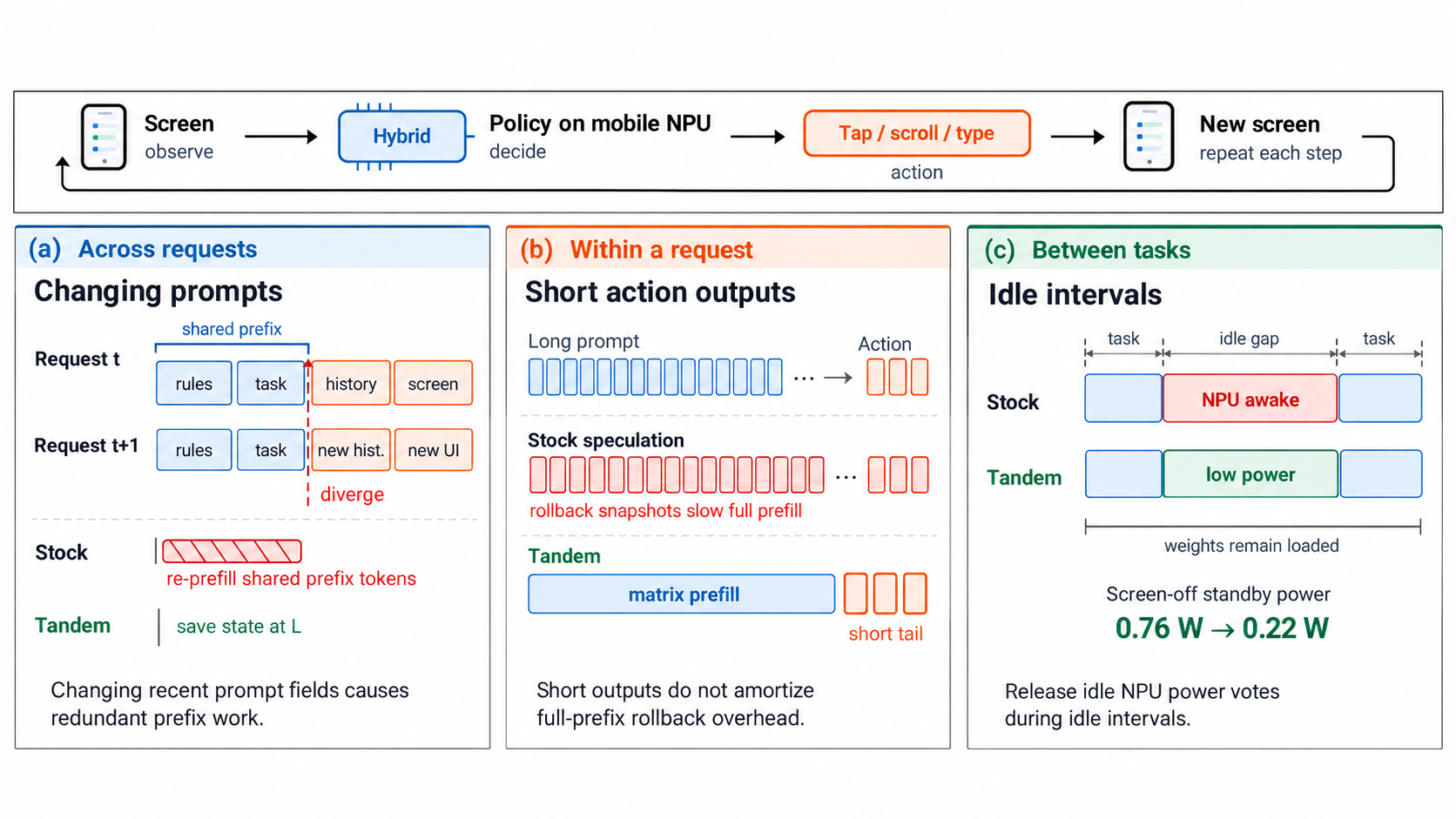}
    \caption{Workload-driven serving challenges for mobile GUI agents. Each step observes the screen and generates an action. (a) Consecutive prompts share a prefix but diverge in recent history or screen content, making chat-oriented checkpoints poorly placed. (b) Short outputs provide few opportunities to amortize the prefill overhead of speculative rollback states. \sys retains matrix-based prefill and uses a short recurrent tail. (c) Between tasks, active NPU power requests can waste energy even when the model remains idle. \sys releases these requests without unloading the model.}
    \label{fig:teaser}
\end{figure*}

Consecutive agent requests share instructions and task information but rewrite recent action history and the screen description. Checkpoints placed at message boundaries or near a prompt's end prepare a server for chat continuation. In an agent request, the next change can occur earlier, within the history or screen description. A recurrent state has already absorbed all tokens up to its saved position: a checkpoint after the change cannot be reused, while one saved too early leaves shared input to process again. 
\zm{In our replay trace, the stock checkpoint policy reprocesses an average of 111 shared tokens per subsequent request (\secref{sec:char-reuse}).}
Therefore, reuse depends on placing checkpoints where the agent's requests actually diverge.

Short action outputs create a second mismatch. Speculative decoding proposes several tokens for the model to check together~\cite{gloeckle2024multitoken,leviathan2023speculative}. Rejecting a proposal requires restoring an earlier recurrent state. In llama.cpp's existing Hexagon path~\cite{ggml2023llamacpp,ggml2026hexagon}, retaining these rollback states moves prompt processing from the matrix kernel to per-token execution. This adds an input-processing cost that a short answer has few decoding rounds to recover. In a controlled ablation, enabling the model's multi-token prediction (MTP) head with the existing rollback and verification paths adds 0.67~s per text step (\secref{sec:char-spec}). Most of the increase occurs before output generation begins.
% src: sec:char-reuse (make_figdata.py, requests 2-61 of the replay trace: 538.5 vs 427.9 tokens per step, gap 111);
%      sec:char-spec (+666 ms, results/abl5 block 1 on the final build, screen verifiably on); sec:char-energy (IDLE_POWER.md sec. 11.1: screen
%      on 0.48 -> 0.89 W, 1.85x; screen off 0.25 -> 0.76 W, 3.0x; 0.41 / 0.51 W x 300 s = 123-153 J = 11-14 text steps)

Prior mobile inference systems accelerate prompt processing, divide work across processors or speed up output generation~\cite{xu2025llmnpu,chen2025heterollm,xue2024powerinfer2,wei2025tman,yin2026shadownpu}. 
% Their full-attention models retain separate cache entries for past tokens, allowing prefix-reuse systems to truncate the cache at the shared input~\cite{zhang2026vlmcache}. 
\zm{These systems largely target full-attention models, whose KV caches allow prefix-reuse systems to truncate the cache at the shared input~\cite{zhang2026vlmcache}.}
% Server systems manage recurrent states with checkpoints where requests branch~\cite{pan2025marconi} and GPU kernels for speculative verification~\cite{wang2024mamballama,wu2025stree}. On-device agent acceleration also addresses prompt reuse and draft generation~\cite{chung2026agentx}. For a hybrid agent on a phone NPU, the remaining challenge is to retain useful states from one changing request stream while preserving matrix execution for prompt processing and sharing weight traffic across short verification batches.
\zm{Prior work studies recurrent-state checkpointing for hybrid models~\cite{pan2025marconi,shirokikh2026sparse} and GPU-based speculative verification~\cite{wang2024mamballama,wu2025stree}. On-device agent acceleration also addresses prompt reuse and draft generation~\cite{chung2026agentx}. Serving hybrid GUI agents on mobile NPUs brings these requirements together: recurrent checkpoints must remain reusable as prompts change, and speculative rollback must not sacrifice efficient matrix-based prefill. Keeping a model resident between tasks also introduces an idle-power cost.}

% \begin{figure*}[t]
%     \centering    
%     \includegraphics[width=\linewidth]{figures/figure1.pdf}
%     \caption{\sys serves GUI agents on a phone's neural processing unit (NPU). (a)~A task advances through repeated screen observations and actions. (b)~The hybrid model combines recurrent layers (loops) with attention (grids). (c)~\sys reuses shared context, reduces token-generation work and releases idle compute resources (\secref{sec:reuse}, \secref{sec:decode}, \secref{sec:power}).}
%     \Description{Three-panel illustration. Left: a smartphone inside a loop of three icons, a screenshot, a neural network
%     and a tapping finger. Middle: a column of layer icons, three circular arrows followed by a square grid, repeated, above
%     an NPU chip. Right: three rows, each an icon for a serving cost, an orange arrow and an icon for the remedy: recomputed
%     pages become a bookmark on the line where two pages differ, crossed-out branches with an undo arrow become a
%     fast-forward arrow, and a draining battery with a clock becomes a power button on the chip.}
%     \label{fig:teaser}
% \end{figure*}

We present \sys, a runtime for hybrid GUI agents on mobile NPUs (Figure~\ref{fig:teaser}). 
% Its key idea is to choose state boundaries according to two reuse needs: shared context across requests and the recent token positions needed within a speculative round. 
% \zm{Its key idea is to manage recurrent state according to two distinct needs: retaining reusable checkpoints across requests and preserving recent states for rollback within speculative rounds.}
For cross-request reuse, the point where consecutive prompts diverge predicts a prefix that later steps can reuse while instructions and task text remain fixed. 
% \sys saves state at that point and consolidates the remaining input batches to remove extra model passes used for chat continuation.
\zm{\sys saves state at that point and consolidates input batches to avoid extra model passes introduced by chat-oriented checkpointing.}

For speculative generation, only a short window of recent token positions needs rollback states. \sys keeps most prompt processing on the matrix unit and uses a per-token tail to produce these states. A column-shared verification kernel reads each weight block once for all drafts, and a trimmed draft vocabulary lowers proposal cost. Fused operators reduce intermediate memory traffic. Between request bursts, \sys retains model weights but releases the NPU session's active power requests, which would otherwise keep the accelerator awake. The next task restores active execution without reloading the model.

We implement \sys in llama.cpp's server and Hexagon backend. On a Snapdragon 8 Elite, it reduces text-step latency by 30\%, screenshot-step latency by 16\% and energy per text step by 33\% relative to llama.cpp's fastest NPU configuration. Standby power with the model loaded falls from 0.76 to 0.22~W. On the next Snapdragon generation, the text-step latency reduction is 27\%. Our evaluation also examines model sizes, other recurrent architectures and the action outputs produced under each optimization.
% src: figures/comparison.tex (see abstract.tex); tab:model; results/abl5 (+M +666 ms; PCGWMHRDV -203 ms vs PCGWV);
%      IDLE_POWER.md sec. 11 (screen off 0.76 -> 0.22 W; first request after a 30 s pause +21.0 ms [+8.8, +33.0]);
%      results/screen.log (exp_screen.sh arch, 2026-10-02); results/dev2/dev2_tail.log (v81: 1.212 vs 1.655 s, 26.8%);
%      energy: tab:energy (32.9%)

This paper makes three contributions.
\begin{itemize}
\item A characterization of hybrid GUI-agent inference on a phone NPU, identifying how checkpoint placement, speculative rollback and active NPU sessions interact with changing prompts, short outputs and idle intervals (\secref{sec:char}).
\item The design of \sys, with divergence-point checkpoints for context reuse, rollback-aware prefill and column-shared verification for speculative execution, and idle power release while retaining the model (\secref{sec:design}).
\item An implementation and evaluation across agent workloads, model sizes, hybrid architectures and two NPU generations, measuring latency, energy and action fidelity (\secref{sec:eval}).
\end{itemize}

% Background, Route 1 draft revision 2 (2026-09-29). Plan: ROUTE1_OUTLINE.md sec. 5 row 2 (1.0 page).
% Sources: research/route1/CHARACTERIZATION.md (C) sec. 1-2 and 4.2, research/npu/NOTES.md (N) sec. 1-3, 8 and 11,
% research/route1/IDLE_POWER.md sec. 2, research/route1/TAILCKPT.md sec. 1. Build ids appear only in src comments.
\section{Background}\label{sec:background}

\subsection{The agent step}\label{sec:bg-step}

The agent executes a task as an \emph{episode} of screen observations and actions. It reads Android's accessibility tree, asks the model for an action, executes it and waits for the app to update. Each action determines the next screen, so model requests form a sequential chain. Model latency accumulates across steps, while app interaction and pauses between tasks leave intervals with no inference work.

Each request contains fixed rules, the task and foreground app, recent actions and a text description of the on-screen controls. The model returns a short JSON response specifying an action, such as tapping an indexed control, entering text or finishing the task. When accessibility data is sparse, the app also attaches a screenshot. Our policy is Qwen3.5-4B~\cite{qwen2026qwen35} fine-tuned with LoRA~\cite{hu2022lora} on public GUI-agent data and demonstrations recorded on the phone.

% Between steps, the app rebuilds this message from the current observation and the last three actions. For example, after opening a settings page, the next request carries the new controls and a record of the preceding action, while retaining the rules and task. Once the history window is full, its oldest entry disappears on the next step. The resulting prompts overlap but also replace text from the previous prompt. The server can exploit this overlap even though the client sends each request independently. It identifies where the reusable prefix ends before advancing the model through the new screen.
\zm{Between steps, the app rebuilds this message from the current observation and the last three actions. For example, after opening a settings page, the next request carries the new controls and a record of the preceding action, while retaining the rules and task. Once the history window is full, its oldest entry disappears on the next step. The resulting prompts overlap but also replace text from the previous prompt.}
% src: research/data/local_format.py, state() and messages(); TEXT_REUSE.md sec. 1.

Serving a request has two phases. \emph{Prefill} processes the input tokens to build the model's state; \emph{decode} produces the action one token at a time. Reusing state from a shared prompt prefix skips part of prefill. Speculative decoding lets a lightweight drafter propose tokens that the full model verifies together, reducing decode passes when several proposals are accepted.

Screenshots add a vision encoder that converts pixels into feature vectors. These vectors form image tokens in the language model's input, adding prefill work before action generation. Reducing the image-token budget changes both this work and the visual information available to the policy.
% src: IMAGE_STEP.md sec. 1 and 5; image-budget measurements in sec:eval-image.

% src: research/data/local_format.py (one user message, enable_thinking false, 3 history entries, 55 controls, action
%      vocabulary in RULES); C sec. 1 (prompt format), sec. 2 (309-token common prefix, single format), T2.1 (answer +
%      EOS 13.9-14.5 tokens); sections/method.tex (screenshot rules from Agent.java; LoRA recipe)

\subsection{Hybrid linear-attention models}\label{sec:bg-hybrid}

Hybrid models combine two ways of retaining context. Attention layers keep separate entries for past tokens in a key-value (KV) cache. Recurrent layers repeatedly update one compact state, folding each new token into the information already stored. A \emph{checkpoint} is a copy of this state saved at a particular token position so that execution can resume there later.

Consider two prompts that share their first 350 tokens and differ afterward. Once the first prompt has been processed, its KV cache can be truncated to 350 entries. Its final recurrent state, however, also contains the later tokens; the server needs the state saved after token 350 to resume there. If its latest usable copy is at token 300, it reprocesses 50 shared tokens before reaching the new suffix. A hybrid model performs both operations: it restores the recurrent checkpoint and aligns the attention cache to that position.
% src: illustrative token counts; restore semantics from CHARACTERIZATION.md sec. 4.2 and TEXT_REUSE.md sec. 1.

Qwen3.5-4B has 32 layers~\cite{qwen2026qwen35}. Every fourth layer is a full-attention layer, and the other 24 are Gated DeltaNet layers~\cite{yang2025gated}. Each of the 32 heads of a Gated DeltaNet layer keeps a matrix state $S$ and updates it once per token,
\begin{equation}
S_t = \alpha_t\, S_{t-1}\bigl(I - \beta_t\, k_t k_t^{\top}\bigr) + \beta_t\, v_t k_t^{\top}, \qquad o_t = S_t\, q_t ,
\label{eq:gdn}
\end{equation}
where $q_t$, $k_t$ and $v_t$ are the token's query, key and value, $\alpha_t$ is a decay gate and $\beta_t$ a write strength. A short causal convolution over the layer's inputs precedes the update. The recurrent state of all 24 layers occupies 50.25\,MiB per sequence regardless of prompt length; the 8 attention layers add 32\,KiB of KV cache per token. Prefill evaluates Eq.~\eqref{eq:gdn} in chunks with matrix products~\cite{yang2025gated}, while decode updates the state once per generated token. These two execution forms become important when the server needs intermediate states for rollback.
% src: C sec. 1 (GGUF metadata: full_attention_interval 4, 4 KV heads of dim 256, 32 v-heads, state 128, conv kernel 4)
%      and T4.2 (2 MiB + 96 KiB per GDN layer, 50.25 MiB, 32 KiB/token of attention KV, equal size at 1,608 tokens)

The model's multi-token prediction (MTP) head~\cite{gloeckle2024multitoken} provides the lightweight drafter. It uses information already computed by the main model to propose the next token, then repeats this process to form a short sequence. The main model checks the sequence in one batched pass~\cite{leviathan2023speculative}. Accepting several proposals can replace several ordinary decode passes. If a proposal is rejected, execution resumes from the last accepted token, requiring both the attention cache and the recurrent state at that position.
% src: C T4.2 (MTP nextn KV 4 KiB/token); N sec. 3

\subsection{The Hexagon NPU}\label{sec:bg-npu}

The Hexagon NPU provides two execution paths used by the model: a matrix unit (HMX) for processing many token vectors together and vector units (HVX) for smaller operations. On the Snapdragon 8 Elite, Hexagon v79 has six hardware threads with 1024-bit HVX units, an HMX unit for fp16 tiles and 8\,MB of tightly coupled memory (VTCM). This local memory holds data close to the compute units; DMA engines transfer larger blocks between it and DRAM. Model weights exceed the local capacity, so execution repeatedly loads weight blocks and applies them to the available tokens. The amount of useful work per transfer determines how well computation can keep up with this data movement.
% src: N header (6 HVX threads, 1 HMX, 8 MB VTCM), sec. 2 (scalar loads from VTCM very slow on v79), negative results
%      (in-order DSP stalls on DDR misses), sec. 8 (DMA 57.8 GB/s by default, 63.7 GB/s with DDR performance mode)

The amount of work available at once determines how these units are used. During prefill, a weight matrix is applied to many token vectors, allowing HMX to reuse its tiles across the batch. Ordinary decode presents one token vector at a time, and quantized matrix--vector products run on HVX. Speculative verification supplies a few vectors together. This creates an opportunity to share weight traffic across drafts, provided the kernel processes those vectors jointly. The relevant batch size is the number of tokens in a model pass, including tokens from a single request.
% src: research/npu/NOTES.md sec. 1-3; the prefill and small-batch paths of the Hexagon backend.

The host communicates with the NPU through FastRPC. llama.cpp's Hexagon backend~\cite{ggml2026hexagon} loads a DSP library, maps shared tensor buffers and submits batches of operators to a work queue. The session also issues \emph{power votes}, requests to the platform to keep compute and memory resources at active performance levels. These votes hold clocks high and prevent low-power modes until the session ends. A loaded model can therefore keep the accelerator powered even when no operators are running.
% src: IDLE_POWER.md sec. 2 rows 1-5 (votes of htp_iface_start: DCVS off, min = target = max corners, sleep_disable =
%      LPM level 1, HMX on at the top clock); N sec. 11 (DSP work queue)

% \subsection{How serving stacks keep hybrid state}\label{sec:bg-serving}
\subsection{\zm{Hybrid-state management in llama.cpp}}\label{sec:bg-serving}

llama.cpp~\cite{ggml2023llamacpp} runs all layers of the hybrid model on the Hexagon NPU. Its server compares each incoming prompt with the input it processed last and finds their longest common prefix. For an attention-only model, it truncates the KV cache at that position and processes the new suffix. For a hybrid model, it restores the newest saved recurrent state at or before that position and aligns the attention cache with it. Any shared tokens between the saved state and the end of the common prefix are processed again.

Checkpoint placement determines how much of this repeated work is necessary. 
% The stock server saves state at user-message boundaries and near the end of a prompt to support chat continuation. Agent requests can change earlier, within the action history or screen description, so these saved positions may not match the next request's reusable input. Each checkpoint also ends an input batch and copies about 50\,MiB of recurrent state.
\zm{The stock server saves state at user-message boundaries and near the end of a prompt to support chat continuation. Each checkpoint also ends an input batch and copies about 50 MiB of recurrent state.}
Saving more positions therefore consumes memory and adds execution passes. A \emph{slot} is the server context that holds a request's cached tokens and model state; it retains a bounded set of checkpoints between requests. A host prompt cache can reload states from earlier requests with a longer matching prefix.
% src: common/common.h n_ctx_checkpoints = 32 (b11200); both engines run with --ctx-checkpoints 8.
% src: C sec. 4.2 (restore, creation, eviction and prompt-cache rules with server-context.cpp / server-task.cpp lines;
%      a checkpoint holds the recurrent row and the MTP KV, not the attention KV); TAILCKPT.md sec. 1 (checkpoint_offsets
%      {4 + n_ubatch, 4}, PR 20288, generation prompt of Qwen3.5 with thinking off)

Speculative decoding also saves states within a request. Verification updates the recurrent state as it checks each proposed token. If only the first two of three proposals are accepted, for example, the server needs the state after the second proposal to continue. These short-lived rollback states serve a different purpose from checkpoints retained for future requests. The backend produces them at batch boundaries, including during prompt processing; \secref{sec:char-spec} examines the cost of doing so on the NPU.
% src: N sec. 3 (K = n_rs_seq + 1 = 4 snapshots per ubatch; the trailing n_rs_seq + 1 tokens of a ubatch, so prompt
%      batches need snapshots too)

% Characterization, revision 2, restyled 2026-09-29 to STYLE_MOBISYS.md sec. 5 (findings named by phenomenon; claim,
% numbers, cause, implication; setup in one sentence). The drafting agent's version is kept in the session scratchpad
% (drafts_backup_20260929). Sources: research/route1/WEEK1_RESULTS.md (W), CHARACTERIZATION.md (C),
% results/abl/summary.md (ablation block 1), results/modelctl.log, TEXT_REUSE.md, IMAGE_STEP.md, IDLE_POWER.md,
% research/npu/NOTES.md (N). Figure data: figures/src/make_figdata.py.
% Measurement conditions differ by study: fan-cooled phase breakdown, cooler-based ablation, battery idle power.
\section{Characterizing the agent step}\label{sec:char}

% We profile the fine-tuned, 4-bit Qwen3.5-4B policy on a Snapdragon 8 Elite. The \emph{app set} contains one screen from each of 21 Chinese apps, with a known correct action and both text and screenshot requests. The \emph{replay trace} appends 40 steps from five recorded episodes to these 21 text requests, preserving request order for prefix reuse.
\zm{We profile the fine-tuned, 4-bit Qwen3.5-4B policy on a Snapdragon 8 Elite to understand how GUI-agent requests affect prefill, speculative decoding, and idle power. The \emph{app set} contains one screen from each of 21 Chinese apps, with a known correct action and both text and screenshot requests. The \emph{replay trace} appends 40 steps from five recorded episodes to these 21 text requests, preserving request order for prefix reuse.}

% The phase breakdown uses the app set with fan cooling, \sys's kernels and speculation, and llama.cpp's stock checkpoint and batching policy. The prefix-reuse and speculation studies use the replay trace. Their configurations are specified below; \secref{sec:eval-setup} describes the end-to-end comparisons.
\zm{Figure~\ref{fig:anatomy} and Table~\ref{tab:prompt-composition} report the phase breakdown on the app set with llama.cpp’s stock checkpoint and batching policies, but with \sys’s NPU kernels and speculation enabled under fan cooling. The prefix-reuse and speculation studies instead use the replay trace, with configurations specified in \secref{sec:char-reuse} and \secref{sec:char-spec}. \secref{sec:eval-setup} describes the end-to-end comparisons.}
% src: C sec. 1 (deployed server arguments: 8 checkpoints, 512 MiB prompt cache, up to 3 MTP drafts); unless a
%      finding names a mechanism, the engine runs \sys's kernels under the stock serving policy

\subsection{A step is a medium prefill and a short decode}\label{sec:char-anatomy}

\begin{figure}[t]
\centering
\definecolor{anatText}{HTML}{6a96b7}
\definecolor{anatShot}{HTML}{fbbb6f}
\begin{tikzpicture}
\begin{axis}[
  xbar, scale only axis, width=0.64\columnwidth, height=4.1cm,
  xmin=0, xmax=2.05, ymin=0.4, ymax=6.6,
  ytick={1,2,3,4,5,6},
  yticklabels={Decode,Tail passes,Image prefill,Image encoder,Text prefill,{Cache/copies}},
  xtick={0,0.5,1,1.5,2}, xlabel={time per step (s)},
  tick label style={font=\footnotesize}, label style={font=\footnotesize},
  ytick style={draw=none}, bar width=5pt,
  axis x line*=bottom, axis y line*=left, y axis line style={draw=none},
  xmajorgrids, grid style={black!12},
  title style={font=\small, at={(0,1)}, anchor=south west, yshift=19pt},
  legend style={font=\footnotesize, draw=none, fill=none, at={(0,1.02)},
    anchor=south west, legend columns=2,
    /tikz/every even column/.append style={column sep=7pt}},
  legend cell align=left, area legend,
  point meta=rawx,
  nodes near coords={
    \pgfmathprintnumber[fixed,precision=2,zerofill]{\pgfplotspointmeta}
  },
  nodes near coords align={horizontal},
  every node near coord/.append style={font=\footnotesize, text=black},
]
\addplot[fill=anatText, draw=none, bar shift=4pt] coordinates
  {(0.543,1) (0.088,2) (0,3) (0,4) (0.582,5) (0.096,6)};
\addplot[fill=anatShot, draw=none, bar shift=-4pt] coordinates
  {(0.728,1) (0.284,2) (1.664,3) (1.702,4) (0.922,5) (0.093,6)};
\legend{Text,Screenshot}
\end{axis}
\end{tikzpicture}
\caption{Server time by phase on the app set, with stock checkpoint and
batching policies, \sys's speculative NPU kernels and fan cooling.
Phase times cover text requests 2--21 and screenshot requests 4--21;
totals are 1.31 and 5.39~s.}
% \Description{Grouped horizontal bars compare text and screenshot phase
% times. Text spends 0.58 seconds in prefill and 0.54 in decode.
% Screenshots add 1.70 seconds of image encoding and 1.66 of image-token
% prefill.}
\label{fig:anatomy}
\end{figure}
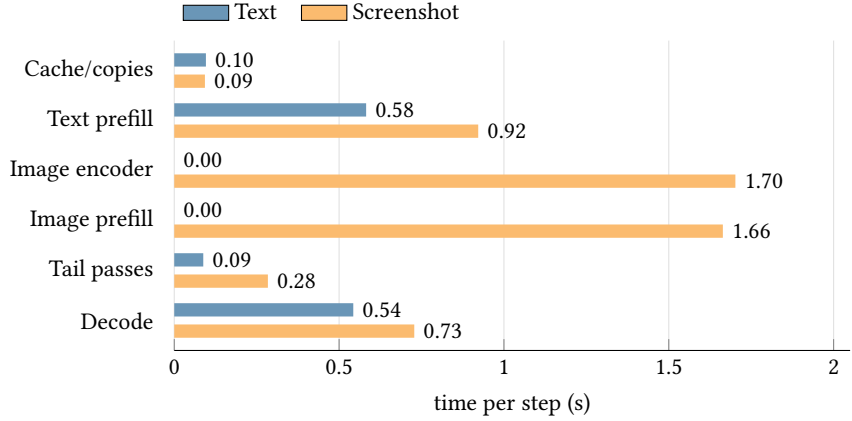

% \begin{figure}[t]
%   \centering
%   \includegraphics[width=0.85\linewidth]{figures/agent_step_phase_times.pdf}
%   \caption{Server time by phase on the app set, with stock checkpoint and
% batching policies, \sys's speculative NPU kernels and fan cooling.
% Phase times cover text requests 2--21 and screenshot requests 4--21;
% totals are 1.31 and 5.39~s.}
%   \label{fig:anatomy}
% \end{figure}

\begin{table}[t]
\centering
\caption{Mean prompt token composition over all 21 screens. The image
count includes two boundary markers.}
\label{tab:prompt-composition}
{\small
\begin{tabular}{@{}lcc@{}}
\toprule
Content & Text & Screenshot \\
\midrule
Rules & 324 & 325 \\
Task and history & 49 & 49 \\
On-screen controls & 275 & 275 \\
Image and markers & 0 & 1,129 \\
\midrule
Total & 648 & 1,778 \\
\bottomrule
\end{tabular}}
\end{table}

An agent step reads far more than it writes. In the app set, a text prompt averages 648 tokens, split mainly between fixed rules and on-screen controls (Table~\ref{tab:prompt-composition}); its answer is about 14 tokens. The controls change with the screen, while the rules provide a reusable prefix.
% src: figures/data/anatomy.dat and figures/src/make_figdata.py (app set: 648 = 324 rules + 49 state/history + 275 controls).
%      For comparison, C T2.1 local_v3 (781.9 tokens; controls 46.9%; rules and wrapper 39.9% + 1.5%; answer + EOS 14.5); C sec. 4.1
%      (consecutive control lists identical in 8/171 and 12/329 pairs)

In the app-set profile, prefill and decode take comparable time: 0.58 and 0.54~s of a 1.31~s text step (Figure~\ref{fig:anatomy}). Decode streams 2.9~GB of weights for every token. Within a prefill pass, element-wise operators and the short convolution account for 30\% of NPU time, largely moving data.

Batch boundaries add a separate cost. A calibrated prefill model estimates 1.30~ms per token plus 86~ms per pass over the model. Ending a batch early therefore costs roughly as much as processing another 66 tokens. The stock policy spends 0.13~s on a four-token tail pass and recurrent-state copies that prepare chat continuation.
\zm{This fixed per-pass cost motivates consolidating unnecessary input batches (\secref{sec:prefill}).}
% src: W T-E3 (single, TAIL=1: cache 49, restore 9, copies 38, tail 88, passes 582, decode 543, step 1310 ms);
%      W T-D1 (1.299 ms/token + 86.5 ms/pass, RMSE 30.2 ms over 16 lengths); N sec. 1 (2.94 GB per token);
%      N sec. 9 correction (448-token pass: matrix 52%, GDN 11%, element-wise + conv 30%)

% A screenshot makes the step \pn{4.1$\times$} as long. The encoder takes \pn{1.70}~s for the 1,127 image tokens and their prefill another \pn{1.66}~s, and the \pn{640} text tokens before the image are recomputed at every screenshot step, because no checkpoint precedes the image.
\zm{Screenshot requests take 4.1$\times$ as long as text requests. The encoder takes 1.70~s to produce 1,127 image tokens, and prefilling those tokens takes another 1.66~s. The stock checkpoint policy also recomputes the 640 text tokens before the image at every step because it retains no checkpoint before the image. The image-related work limits the gains available from text-prefix reuse alone.}
% src: W T-IMG row "switch unset" (text 922 ms for 640 tokens, encode 1702, image tokens 1664, step 5398 ms);
%      W sec. 4.3 (R = 0 on 21/21 screenshot steps)

% \subsection{Checkpoints limit prefix reuse}\label{sec:char-reuse}
\subsection{Stock checkpoint placement limits prefix reuse}\label{sec:char-reuse}

% F-REUSE, Route 1 draft revision 2: new prompt tokens per request, hybrid vs full attention. Input from
% sections/characterization.tex.
% data: figures/data/reuse.dat, written by figures/src/make_figdata.py (one row per request of the 61-prompt sequence).
% src: hybrid today = research/route1/results/rt22/ab-rt22-0.jsonl (measured, fresh server, single-message format,
%      upstream checkpoint policy, MTP on; equals the simulator on 61/61 requests); full attention =
%      research/route1/results/mc-rt22-lat.jsonl (Qwen3-VL-4B Q4_0, same engine build, same prompts and order;
%      results/modelctl.log); divergence-point checkpoints = research/route1/tools/textckptsim.py, policy ("exact", 32)
%      = LLAMA_CKPT_DIVERGE=1, SIMULATED (TEXT_REUSE.md sec. 5) and matched by the phone run below.
% src: means over requests 2-21 / 2-61 printed by make_figdata.py: today 460.4 / 538.5, divergence 348.9 / 443.1,
%      full attention 326.0 / 420.0, full-attention bound on the hybrid's tokens 334.2 / 427.9.
% colors: first three validated categorical slots (blue, orange, aqua); secondary encoding by marker and dash pattern.
\begin{figure}[t]
\centering
\definecolor{reuseToday}{HTML}{EB6834}
\definecolor{reuseDiv}{HTML}{2A78D6}
\definecolor{reuseFA}{HTML}{1BAF7A}
\begin{tikzpicture}
\begin{axis}[
  scale only axis, width=0.83\columnwidth, height=3.0cm,
  xmin=0.5, xmax=61.5, ymin=0, ymax=1400,
  xtick={1,11,21,31,41,51,61}, ytick={0,300,600,900,1200},
  tick label style={font=\scriptsize}, label style={font=\scriptsize},
  xlabel={request in the replay trace}, ylabel={new prompt tokens},
  xlabel shift=-3pt, ylabel shift=-4pt,
  axis x line*=bottom, axis y line*=left, ymajorgrids, grid style={black!10},
  legend style={font=\scriptsize, draw=none, fill=none, at={(0.45,1.02)}, anchor=south, legend columns=2,
    row sep=-1pt, /tikz/every even column/.append style={column sep=6pt}},
  legend cell align=left,
]
% zh21 | recorded episodes, and the starts of the five episodes (requests 22, 26, 31, 50, 60)
\draw[black!45, dashed, line width=0.4pt] (axis cs:21.5,0) -- (axis cs:21.5,1400);
\pgfplotsinvokeforeach{25.5,30.5,49.5,59.5}{\draw[black!30, dotted, line width=0.5pt] (axis cs:#1,0) -- (axis cs:#1,1400);}
\node[font=\scriptsize, anchor=north] at (axis cs:11,1390) {app set};
\node[font=\scriptsize, anchor=north] at (axis cs:41.5,1390) {recorded episodes};
\addplot[color=reuseToday, line width=0.5pt, mark=*, mark size=0.9pt] table[x=i, y=today] {figures/data/reuse.dat};
\addplot[color=reuseFA, line width=0.5pt, mark=triangle*, mark size=1.1pt] table[x=i, y=fa] {figures/data/reuse.dat};
\addplot[color=reuseDiv, line width=0.5pt, dashed, mark=square*, mark size=0.75pt, mark options={solid}]
  table[x=i, y=div] {figures/data/reuse.dat};
\legend{{Hybrid, stock checkpoints}, {Full attention}, {Hybrid, divergence-point checkpoints}}
\end{axis}
\end{tikzpicture}
\caption{New prompt tokens per request of the replay trace: the app set, then five recorded episodes (dotted lines mark episode starts). The full-attention Qwen3-VL-4B runs on the same engine. Divergence-point checkpoints (\secref{sec:reuse}) close most of the checkpoint gap.}
% src: the divergence-point series comes from textckptsim.py; the phone run of the rule (results/rt24.run.log,
%      2026-09-29) processed exactly these prompt_n values on 61/61 requests of the trace and 21/21 of the app set
\Description{Line chart over the 61 requests of the replay trace. The orange line for the hybrid model with today's checkpoints lies above the dashed blue line for divergence-point checkpoints on most requests, and the blue line nearly coincides with the aqua line for the full-attention model.}
\label{fig:reuse}
\end{figure}
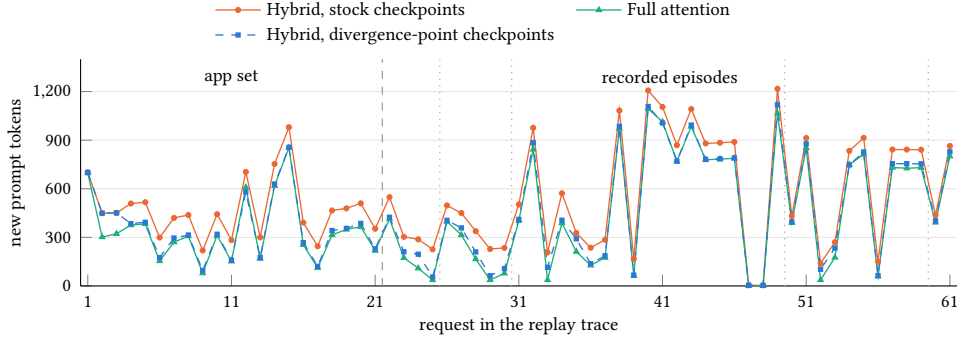

% Consecutive agent prompts share most of their text. 
\zm{Consecutive agent prompts share a substantial text prefix.}
All prompts share the first 309 tokens, and within an episode consecutive prompts diverge a few tokens into the action history, about 360 to 400 tokens in. A KV cache reuses such a prefix exactly. A recurrent state cannot, because it has absorbed every token it has seen and resumes only from a stored checkpoint.
% src: TEXT_REUSE.md sec. 1 (divergence points: task text, foreground app 326-345, history 360-372 and 371-404)

% This difference opens a \emph{checkpoint gap}. Over requests 2--61 of the replay trace, a full-attention Qwen3-VL-4B processes \pn{420} new prompt tokens per step, and the hybrid model \pn{539}, on the same engine (Figure~\ref{fig:reuse}). A KV cache on the hybrid model's own token stream would process \pn{428}. The hybrid therefore recomputes \pn{111} tokens per step, about \pn{140}~ms.
\zm{The checkpoint gap is visible on the hybrid model's own token stream. Over requests 2--61 of the replay trace, the stock policy processes 539 new prompt tokens per step; a KV cache on that same token stream would process 428 (Figure~\ref{fig:reuse}). The difference is 111 recomputed tokens, or about 140~ms per step. For context, the full-attention Qwen3-VL-4B processes 420 new tokens per step on the same engine.}
% src: make_figdata.py and TEXT_REUSE.md sec. 8 (means over requests 2-61 of the replay trace: 420.0 full attention,
%      538.5 hybrid, bound 427.9; gap 110.6 tokens x 1.30 ms = 144 ms; the app set alone, requests 2-21: 326.0 / 460.4 /
%      334.2); modelctl.log; TEXT_REUSE.md sec. 1 (in-episode pairs: gap 58 tokens)

The gap comes from checkpoint placement. For a prompt containing one user message, the checkpoint 516 tokens before the end can fall well before the next request's divergence point; the checkpoint four tokens before the end is usually too late to reuse. Screenshot checkpoints fall inside or after the image, leaving no reusable text checkpoint even when consecutive requests share more than 300 text tokens. The useful checkpoint position follows the overlap between requests, rather than the distance from the prompt's end.
% src: TEXT_REUSE.md sec. 1 and 5.3 (restore at 185 = 701 - 516; L - 185 = 124-141); C sec. 4.3 (a first prompt shorter
%      than 517 or longer than ~826 tokens leaves no usable checkpoint); IMAGE_STEP.md sec. 1 (310-327 shared tokens)

% \subsection{Speculation slows a hybrid model down}\label{sec:char-spec}
\subsection{Stock speculative decoding slows hybrid inference}\label{sec:char-spec}

% In the replay-trace ablation, enabling MTP raises text-step latency from \pn{1.85} to \pn{2.52}~s and decode time from 62 to 67~ms per token. Both configurations use the same prefill and memory-traffic optimizations; speculative execution still uses the stock rollback and verification paths (\secref{sec:eval-ablation}). Most of this slowdown comes from retaining intermediate recurrent states.
\zm{Enabling MTP through the stock rollback and verification paths makes a text step slower, from 1.85 to 2.52~s in the replay-trace ablation. Both configurations use the same prefill and memory-traffic optimizations (\secref{sec:eval-ablation}). Decode time also rises from 62 to 67~ms per token, but most of the step-level increase occurs during prefill, when the engine retains intermediate recurrent states.}
% src: N sec. 1 (2.94 GB per token, ~51 GB/s); results/abl5 (block 1 without log-probabilities, 2026-10-02/03, screen
%      verifiably on; +M: PCGWV 1850 -> PCGWMV 2516 ms, bootstrap [+576, +758]; decode 61.9 -> 66.8 ms/token)

The added prefill cost comes from how the engine prepares to undo a rejected draft. It saves states after the final tokens of every batch, including prompt batches. The chunked Gated DeltaNet kernel on HMX produces only one final state. Requesting intermediate snapshots therefore sends the entire batch through the per-token HVX recurrence. This \emph{rollback tax} adds 614~ms to prefill. A need for a few recent states changes how the whole prompt is processed.

Verification and drafting also add work. The stock vector kernel replicates activations in on-chip memory, making four-token verification 2.1$\times$ as expensive as one token. The draft head then scans a 248,320-row output matrix to propose each new token; vocabulary-sized operations consume 35\% of decode time. Thus both preparing the input and producing the short answer contribute to the slowdown.
% src: N sec. 3 (4 snapshots per sequence; HMX chunked only for one state); results/abl5 (prompt +614 ms);
%      N sec. 2 (q4_K 4096 x 14336: 635 us for 1 column, 1322 us for 4); N sec. 7 (521.5 MB head; 35% of decode)

% \noindent\textbf{Short outputs amplify the prefill cost.} Let $\Delta P$ be the prefill overhead introduced by speculation, $D$ the cost of drafting one token, $V$ the cost of a verification pass, and $G$ the ordinary decode cost per token. If a round drafts $d$ tokens and emits $A$ tokens on average, then for an answer of $N$ tokens the latency difference is approximately
% \begin{equation}
% \Delta T \simeq \Delta P + N\left(\frac{dD+V}{A}-G\right).
% \label{eq:spec-cost}
% \end{equation}
% Here $\Delta P$ is paid once per request, while the term multiplied by $N$ describes the change in output-generation cost. For a 14-token agent answer, the measured 614~ms prefill penalty alone contributes about 44~ms per output token. Even if several drafts are accepted in each round, the short answer offers few rounds over which to recover that initial cost. This motivates preserving fast prefill while reducing the drafting and verification costs, which \secref{sec:decode} addresses together.
\zm{\noindent\textbf{Short outputs amplify the prefill cost.} The rollback penalty is paid once per request, but a short answer offers few decode rounds in which to recover it. For an answer of about 14 tokens, the measured 614~ms prefill penalty is equivalent to roughly 44~ms per output token. Accepting several drafts in one round helps only if the decode saving exceeds this initial cost. Speculation therefore needs both matrix-based prefill and cheaper verification and drafting (\secref{sec:decode}).}
% src: latency decomposition; 614 ms from results/abl5 divided by the approximately 14-token output in CHARACTERIZATION.md T2.1.

\subsection{A loaded model keeps the NPU awake}\label{sec:char-energy}

% A loaded model consumes power between requests.
\zm{A loaded model can leave the NPU powered between requests even when no inference work is running.} 
With the screen on, the idle phone draws 0.48~W without a server and 0.89~W with the model loaded; with the screen off, power rises from 0.25 to 0.76~W. The NPU's power-collapse counter stops advancing, and its core and matrix unit remain at their top clocks at zero utilization. Session-wide power votes cause this \emph{residency tax}, including while the phone is suspended.
% src: W sec. 5.2 (E3 10.5 / 11.4 J per step; E4 0.012-0.0135 J per prompt token, 0.63-0.65 J per generated token);
%      IDLE_POWER.md sec. 11.1 (screen on 0.48 / 0.89 W, screen off 0.25 / 0.76 W, ABBA on battery); sec. 1-2 (frozen
%      collapse counter, core 2,112 MHz and HMX 1,209.6 MHz at 0% utilization)

% The app retains the model for five minutes after a task. The excess idle power over this interval consumes \pn{120--150}~J, equivalent to \pn{12--15} text steps. Unloading the model incurs a \pn{9}~s cold start on the next task. The runtime can instead retain the weights in memory and release the NPU's active power requests during this interval.
\zm{The app retains the model for five minutes after a task. At the measured excess idle power, this interval amounts to 120--150~J, equivalent to 12--15 text steps. Unloading the model reduces that idle draw but incurs a 9~s cold start on the next task. This motivates releasing the NPU's active power requests while retaining the weights in memory.}
% src: 0.41 W (screen on) and 0.51 W (screen off) x 300 s = 123 / 153 J; 10.1 J per text step (tab:energy); docs/local-idle-reclaim-
%      20260929.md (5-minute reclaim; cold start 9.040 s)

% Design, revision 2, restyled 2026-09-29 to STYLE_MOBISYS.md sec. 6 (per mechanism: requirement, why the obvious approach
% fails, key idea, mechanism, payoff with a forward pointer; exactness stated once in the overview). The drafting agent's
% version is kept in the session scratchpad (drafts_backup_20260929). Sources: research/route1/TEXT_REUSE.md,
% IMAGE_STEP.md, TAILCKPT.md, FUSION.md, IDLE_POWER.md, ABLATION_PLAN.md, WEEK1_RESULTS.md (W), results/abl/summary.md
% (ablation block 1: leave-one-out deltas quoted as payoffs), results/rt22/exp_rt22.log; research/npu/NOTES.md (N).
\section{Design}\label{sec:design}

\subsection{Overview}\label{sec:design-overview}

\begin{figure*}[t]
    \centering    
    \includegraphics[width=0.9\linewidth]{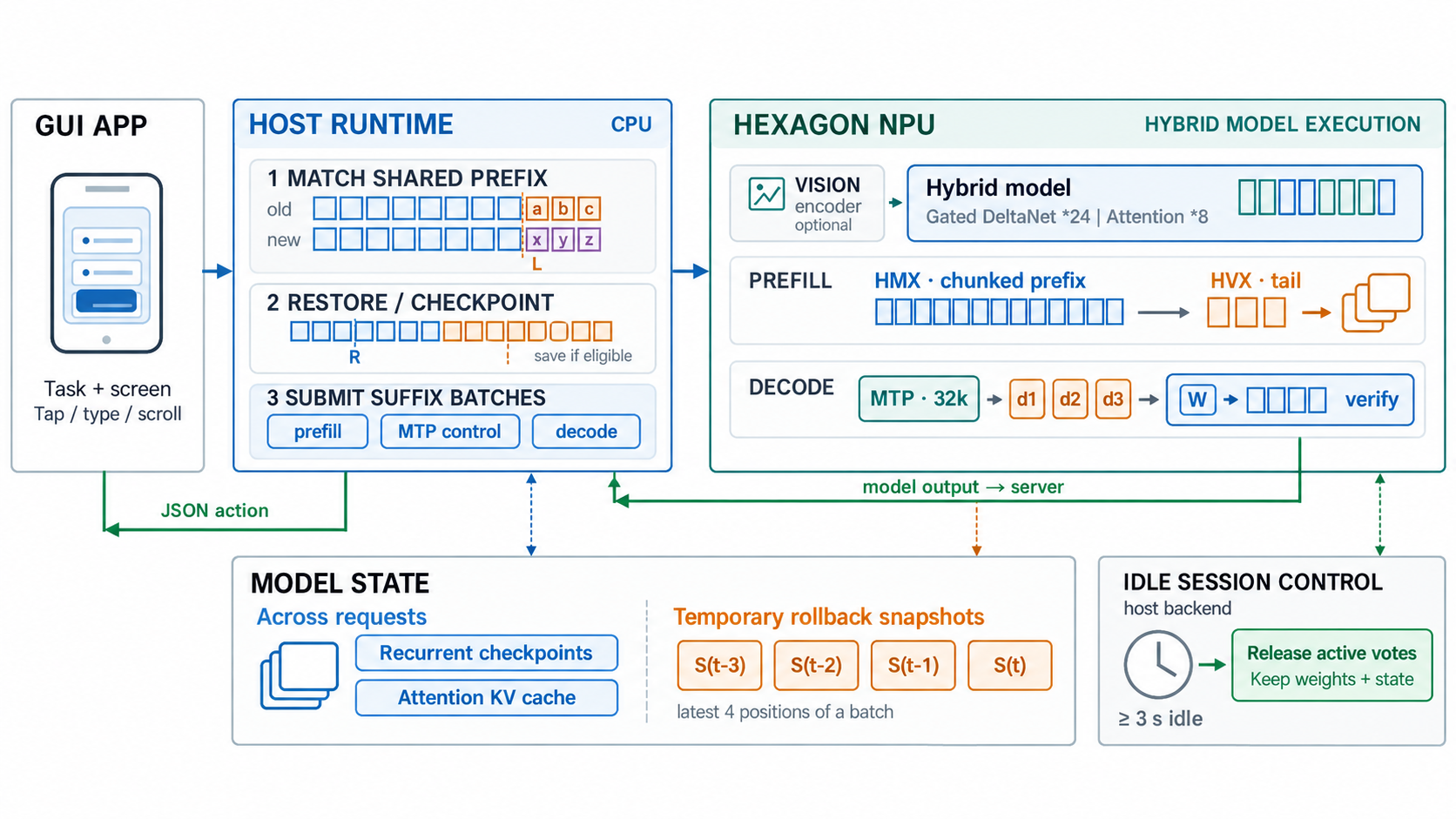}
    \caption{\sys's serving architecture, showing cross-request recurrent checkpoints, temporary rollback states for speculative decoding, and idle NPU session control.}
    \label{fig:overview}
\end{figure*}

\sys follows the agent's request cycle (Figure~\ref{fig:overview}). Across requests, unchanged instructions and task text make the current shared prefix a candidate for reuse by later steps. The server compares the incoming prompt with its cached tokens, restores the latest checkpoint within their common prefix and processes the remaining input. When it reaches the point where the prompts diverge, it can save a new checkpoint for subsequent requests. The placement rule accounts for the cost of creating that checkpoint (\secref{sec:reuse}).

Within a request, speculative verification needs states only for the few recent positions to which execution may roll back. The earlier context needs no per-token snapshots. \sys therefore processes most of each prompt batch on the matrix unit and obtains its remaining rollback states from a short per-token tail. During verification, all proposed tokens share the same weight-block reads. These choices reduce the state-management and verification costs that otherwise outweigh decoding savings on short action outputs (\secref{sec:decode}).

Batch consolidation removes small input passes, while fused operators reduce memory traffic within each pass (\secref{sec:prefill}). For screenshots, the vision encoder supplies new image tokens after the reusable text prefix. The server returns the selected action through the existing interface; the app executes it and supplies the next observation. Between request bursts, the backend releases active NPU resources while retaining model weights and cached state.

% src: results/abl3 (identical answers and first-token top-5 log-probs for P, C, G, W, R, D and I on the replay trace;
%      M, H, T and X change at most one answer per trace, summaries only); R changes the first-token top-5 of screenshot
%      steps under the stock policy: ggml_hexagon_hvx_max_nrows sends m <= 5 (q4_0, q4_1, q4_K, q5_K) and m <= 8 (q6_K)
%      to the HVX rt kernels instead of upstream's m <= 4 (HTP_MM_HMX_MIN_NROWS), RESULTS_20260930.md ("a 5-token ubatch:
%      rt matmul on HVX with int8 activations"); ABLATION_PLAN.md sec. 2-3 (switches)

\subsection{Divergence-point checkpoints}\label{sec:reuse}

Reusing context requires a saved state close to the point where two requests begin to differ. A checkpoint costs a state copy and a batch boundary, so placing one at every potentially reusable token would erase the prefill saving. 
With eight checkpoints per slot in our configuration, the server needs to predict which positions later steps will use. 
% Cloud systems retain branch points in a radix tree of requests~\cite{pan2025marconi}; \sys uses the overlap between consecutive requests from one agent.
\zm{Marconi~\cite{pan2025marconi} also checkpoints recurrent states at branch points in a radix tree. \sys applies a cost-calibrated spacing rule to divergence points observed in one agent's sequential requests.}

Within an episode, rules and task text stay fixed while recent actions and the screen description change. The current divergence point therefore predicts a prefix that the next request can reuse. In the recorded episodes, divergence points remain at or beyond previously retained checkpoints once the episode settles. \sys uses this locality to place checkpoints online.

For each request, let $L$ be the longest common prefix between the new prompt and the tokens cached in the slot, including any entry loaded from the prompt cache. After restoring the newest checkpoint at $R\le L$, \sys splits the batch that crosses $L$ and stores the state there. It does so only when $L$ precedes the prompt's end and any image, and $L-R\ge32$. The rule creates at most one checkpoint per request. Subsequent requests reuse it through the server's existing restore and eviction rules; a request that diverges earlier falls back to an earlier checkpoint. The 32-token threshold avoids closely spaced copies; we select it using a cost model calibrated to the measured prefill and checkpoint costs.

The policy follows changes in tasks and screens. A new task makes requests diverge near the end of the fixed instruction block. The resulting checkpoint can survive later restores within the episode. As the action-history window fills, divergence moves farther into the prompt and may justify a new checkpoint. Once the window starts sliding, requests tend to diverge near its beginning, allowing that position to be reused. These transitions explain why the server creates checkpoints on a minority of requests rather than on every step.
% src: TEXT_REUSE.md sec. 1 and 3.1 (task changes near token 309, history-window filling and steady-state overlap).

Checkpoint spacing trades a cost now for reuse later. 
% With the measured cost of roughly 86~ms per added pass and 1.30~ms per processed token, a checkpoint that advances reuse by 32 tokens needs about three future hits to repay its creation. 
\zm{With a per-pass cost of about 86~ms and a per-token cost of 1.30~ms, advancing reuse by 32 tokens saves about 42~ms per future hit. Roughly three hits would amortize the added pass, excluding the state-copy cost.}
A 64-token threshold misses many of the 25--60-token advances between prompt fields. We sweep the threshold using a replay model of the server's restore, eviction and prompt-cache rules. The model reproduces 422/422 measured prompt-token counts; under its calibrated latency model, the 32-token setting is within 4~ms of the best setting on each trace.
% src: TEXT_REUSE.md sec. 3.3 and 5.1. Threshold latency comparisons are model estimates, not additional timed runs.

On the replay trace, the rule creates a checkpoint on 10 of the 60 requests after the first and retains one additional state copy of about 50\,MiB. Over these 60 requests, new prompt work falls from 539 to 443 tokens per step, within 15 tokens of the KV-cache reuse bound. The same rule applies to the text prefix of screenshot requests.
% src: TEXT_REUSE.md sec. 2 (LLAMA_CKPT_DIVERGE=1: n_lcp before the restore; break at L if L < n_text and L >= n_past +
%      32), sec. 3.1 (the 309-token checkpoint survives), sec. 3.2 (at most one checkpoint per request), sec. 3.3 (32 within
%      4 ms of the best gap); make_figdata.py and sec. 8 (requests 2-61 of the replay trace: 443.1 vs 538.5 vs bound
%      427.9; app set 348.9 / 460.4 / 334.2; creating steps 10/60, app set 2/20; one more retained checkpoint); phone run
%      results/rt24.run.log (prompt_n = simulation on every step); results/abl5 leave-one-out -X +79 ms (final build)

\subsection{Rollback-aware speculative decoding}\label{sec:decode}

Speculation can generate an action with fewer model passes. On this NPU, that saving depends on preparing rollback states without slowing the whole prompt and checking several drafts with less work than separate decode passes. \sys changes the prefill path, the verification kernel and the draft head to address these costs together.

% Execution schematic derived from research/npu/NOTES.md sec. 3 and gated_delta_net_f32_hybrid.
% n=448 and K=4 are an illustrative batch, not a new timing measurement. Widths are schematic.
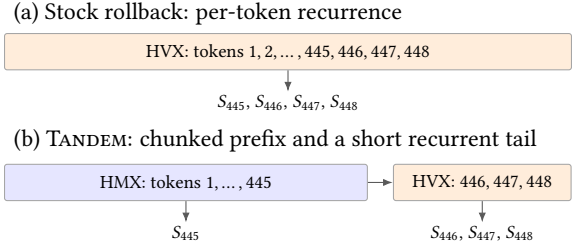
\begin{figure}[t]
\centering
\begin{tikzpicture}[x=1cm,y=1cm,
  box/.style={draw=black!35,rounded corners=1pt,minimum height=0.48cm,font=\scriptsize,inner sep=3pt},
  state/.style={font=\scriptsize,inner sep=1pt},
  arrow/.style={-{Latex[length=1.3mm]},thin,black!65}]
\node[anchor=west,font=\small] at (0,2.35) {(a) Stock rollback: per-token recurrence};
\node[box,fill=orange!14,minimum width=7.5cm] (stock) at (3.75,1.85)
  {HVX: tokens $1,2,\ldots,445,446,447,448$};
\node[state] (stockstates) at (3.75,1.18) {$S_{445}$, $S_{446}$, $S_{447}$, $S_{448}$};
\draw[arrow] (stock.south) -- (stockstates.north);
\node[anchor=west,font=\small] at (0,0.68) {(b) \sys: chunked prefix and a short recurrent tail};
\node[box,fill=blue!10,minimum width=4.8cm] (bulk) at (2.4,0.12)
  {HMX: tokens $1,\ldots,445$};
\node[box,fill=orange!14,minimum width=2.35cm] (tail) at (6.32,0.12)
  {HVX: $446,447,448$};
\draw[arrow] (bulk.east) -- (tail.west);
\node[state] (first) at (2.4,-0.55) {$S_{445}$};
\node[state] (rest) at (6.32,-0.55) {$S_{446}$, $S_{447}$, $S_{448}$};
\draw[arrow] (bulk.south) -- (first.north);
\draw[arrow] (tail.south) -- (rest.north);
\end{tikzpicture}
\caption{Producing four rollback states for a 448-token batch. The stock path runs the entire recurrence on HVX. \sys obtains the first required state from HMX and uses HVX for the three remaining updates. Widths are schematic.}
\Description{Two execution paths produce the recurrent states after tokens 445 through 448. The stock path uses HVX for all 448 tokens. Tandem uses HMX for the first 445 tokens and HVX for the final three.}
\label{fig:rollback}
\end{figure}

\noindent\textbf{Rollback-aware recurrent prefill.} Undoing rejected drafts requires only the states near the end of a batch. GPU verification kernels recompute or scan recurrent states for candidate tokens~\cite{wang2024mamballama,wu2025stree}. On Hexagon, requesting those states displaces the chunked HMX kernel even during prefill. \sys divides the batch into a large prefix that uses HMX and a short tail that saves each intermediate state.

With up to three drafts, the engine retains the final $K=4$ states of a batch. For $n\ge K$ tokens, \sys processes the first $n-K+1$ tokens with the chunked kernel and writes the resulting state directly into snapshot slot $K-1$. The HVX recurrence continues from that state for the remaining $K-1$ tokens, saving each intermediate state. Shorter batches use the per-token path. All required snapshots are available, while all but three tokens of a large batch use the matrix unit.
% Figure~\ref{fig:rollback} illustrates a 448-token batch. The chunked prefix ends at token 445, producing the first required state; three vector updates produce the remaining states. The vector tail stays at $K-1$ tokens as the prompt batch grows.
\zm{Figure~\ref{fig:rollback} illustrates this division for a 448-token batch.}

% Context checkpoints preserve shared prefixes for later requests (\secref{sec:reuse}). Rollback states preserve the recent token positions within one request, allowing verification to restore the state at the last accepted token. The server retains context checkpoints across requests and refreshes rollback states at each batch.

% src: N sec. 3 (apply_gdn_hybrid.py: chunked HMX over [0, n-K+1) into slot K-1, HVX over the last K-1 tokens);
%      results/abl5 leave-one-out -H: +460 ms text [+385, +537], +1422 ms screenshot

\noindent\textbf{Column-shared verification kernel.} A verification pass applies the same weights to the last accepted token and up to three drafts. Processing these token vectors together lets them share weight reads. \sys dequantizes each weight block once, and HVX threads apply it to all token columns with vector-by-scalar instructions before advancing to the next block. Four-column verification takes 1.27$\times$ the time of one column, versus 2.1$\times$ with the stock kernel.

The activation quantizer writes compact int8 values to DRAM, packing four values into each scalar-register load. This removes the stock kernel's 32-fold activation replication and avoids slow scalar reads from VTCM. Per-block activation sums account for weight-format offsets. The backend uses this kernel for products with up to five columns, or eight with 6-bit weights; larger products use HMX.
% src: research/npu/NOTES.md sec. 2 (weight-block reuse, scalar int8 activation groups, stock replication and routing).
% src: N sec. 2 (vrmpy(Vu.ub, Rt.b); 512 KB DDR scratch; bias correction for q4_0 and q6_K; HVX <= 5 rows for
%      q4_0/q4_K/q5_K, <= 8 for q6_K; q4_K 4096 x 14336: 623 / 789 us for 1 / 4 columns, stock 635 / 1322 us);
%      results/abl5 leave-one-out -R: +176 ms [+142, +213]

\noindent\textbf{Trimmed draft head.} Proposing a token should cost substantially less than generating it with the full model. The MTP head uses a lightweight transformer block, but its 248,320-row output matrix still occupies 521.5\,MB. Multiplying by that matrix and selecting a token costs about 12~ms per draft. Following frequency-ranked drafting~\cite{zhao2025frspec}, \sys copies 32,768 rows into a 68.8\,MB draft matrix: special and byte tokens, the first 8,000 CJK tokens and frequent tokens from the agent's training data.

The model file also stores each selected row's original token ID. The draft sampler maps its chosen row back to that ID before verification, so the compact matrix integrates with the existing token stream. Verification and the final output decision use the target model's full vocabulary.
% src: N sec. 7 (9.9 ms per draft token for the q6_K head + 2 ms top-k; 32,768 rows of token_embd, 68.8 MB);
%      results/abl5 leave-one-out -D: +136 ms [+109, +164]

The speculative path adds 201\,MiB for four recurrent-state planes and 68.8\,MB for the draft matrix. \secref{sec:eval-ablation} separates its prefill, verification and drafting costs.
% src: results/abl5 (PCGWMHRDV 1647 ms vs PCGWV 1850 ms, PCGWMV 2516 ms); C sec. 4.4 (RS buffer with 4 rollback planes, 201 MiB)

\subsection{Lean prefill}\label{sec:prefill}

After context reuse removes shared input, the remaining prompt still passes through every model layer. Each additional batch repeats that traversal, and separate operators within a layer may write intermediate results to DRAM only to read them back immediately. \sys reduces these costs by consolidating input batches and keeping intermediate data close to the NPU's compute units.

\noindent\textbf{Batch consolidation.} The stock checkpoint policy ends batches 516 and 4 tokens before the prompt's end, preparing a state for chat continuation. This leaves a four-token pass even when those tokens would fit in the preceding batch. For the agent's independently submitted requests, \sys replaces these breaks with one 512 tokens before the end, matching the server's micro-batch limit. After restoring a checkpoint, the server updates its eviction order and reuses the saved copy.

% For an uninterrupted suffix of 500 tokens, the stock policy runs 496 tokens and then the final four through separate model passes. Consolidation processes all 500 together. With 600 tokens remaining, the stock boundaries produce passes of 84, 512 and 4 tokens; the new boundary produces 88 and 512. The token sequence is the same, but one traversal of the model and its associated checkpoint copy disappear. Shifting the earlier boundary is necessary: removing only the final break would leave 516 tokens, which the 512-token micro-batch limit would split again. On the replay trace, consolidation raises new prompt work from 443 to 460 tokens per step but removes one pass from every step.
\zm{With 600 tokens remaining, the stock boundaries produce passes of 84, 512 and 4 tokens; the new boundary produces 88 and 512. The token sequence is unchanged, but one model traversal and its associated checkpoint copy disappear. Shifting the earlier boundary is necessary: removing only the final break would leave 516 tokens, which the 512-token micro-batch limit would split again. On the replay trace, consolidation raises new prompt work from 443 to 460 tokens per step but removes one pass from every step.}
% src: illustrative suffix lengths under the batch-boundary rules in TAILCKPT.md sec. 1-2. Replay trace, requests 2-61:
%      443.1 new prompt tokens per step with divergence-point checkpoints alone (nl-main-T*, 2026-10-01), 459.9 with
%      batch consolidation as well (sc-main-T*, 2026-10-02); net effect in the ablation: -T +170 ms (results/abl5).
% src: TAILCKPT.md sec. 1-2 (offsets {n_ubatch}; re-snapshot of the restored state); W T-E1/T-E2 (decode +24-31 ms per
%      request with the rule; net effect in ablation block 1 not significant: -T -25 ms [-71, +14]) -> evaluation

\noindent\textbf{Fused SwiGLU.} A feed-forward block multiplies its input by gate and up matrices and combines the products with SwiGLU. The stock backend writes these temporary products to DRAM and reads them back in a separate operator. \sys computes paired columns of the two matrices in adjacent VTCM tiles. While HMX processes the next chunk, HVX applies SwiGLU to the completed tiles and stores only the combined output. At 448 tokens, this removes 66\,MB of traffic per layer while preserving the arithmetic order.

% src: FUSION.md sec. 1-4; research/npu/NOTES.md sec. 9-10.
% src: FUSION.md sec. 1-4 (loop order, VTCM plan, exactness argument; simulator test over all 65,536 fp16 values and
%      1,047 shapes, 3,785,992 elements, 0 mismatches), sec. 6 (fusion conditions); W (464 -> 438 ms NPU prefill)

Pairing gate and up weights within one HMX job retains the pipeline's two output buffers. Alternating separate jobs would require four buffers, exceeding the VTCM budget for a 448-token batch. The host enables fusion when the plan fits the session's memory budget without adding row blocks or changing the pipeline mode; otherwise it keeps the separate operators.
% src: FUSION.md sec. 3 (alternative schedules, VTCM budget and host eligibility checks).

\noindent\textbf{Channel-major convolution.} The recurrent layers' short convolution operates along time within each channel. \sys keeps this channel-major layout, using 2D DMA to stream channel slices into VTCM for convolution and the following SiLU. This removes the stock path's two transposes, each 14.7\,MB per layer at 448 tokens. During decode, one DMA per thread and vector gathers move the short per-channel state rows. The backend also requests the memory controller's performance mode during active execution.
% src: N sec. 10 (fused CONCAT + SSM_CONV + SILU; two transposes of 14.7 MB per layer), sec. 6 (decode-time CONCAT
%      3.17 -> 0.32 ms/token, CPY 1.35 -> 0.51 ms/token)

% src: N sec. 8 (DMA 57.8 -> 63.7 GB/s); results/abl5 cumulative +P -20, +C -37, +G -28, +W -42 ms (1977 -> 1850 ms);
%      leave-one-out in the final configuration -P 0, -C +41, -G +14, -W +18 ms

\subsection{Image path}\label{sec:image}

% A screenshot introduces new visual content even when much of the accompanying text is unchanged. \sys handles the two parts separately: it reuses the processed text prefix and runs the new image through the NPU vision encoder. The encoder takes about 1.7~s per screenshot and supplies image tokens to the language model. Keeping it on the NPU avoids a lengthy CPU stage before the model can process the screen.
\zm{\sys reuses the processed text prefix and runs the new screenshot through the NPU vision encoder. Keeping the encoder on the NPU avoids a lengthy CPU stage before the model processes the screen.}
% src: IMAGE_STEP.md sec. 5 (all vision ops on HTP0 incl. FLASH_ATTN_EXT; UPSCALE on the CPU, 27.9 ms = 1.7%);
%      docs/local-model-perf-2026-09-27.md sec. 1 (encoder GPU 13-22 s, NPU 1.7 s); results/abl5 -V (screenshot prompt
%      54.6 s with the encoder on the CPU, 4 threads, vs 4.1 s on the NPU: +50.5 s, so about 52 s of CPU encoding)

The divergence rule checkpoints text before the image, where multimodal rotary positions still equal token indices. Restoring this state reuses the shared text before encoding and prefilling the new screenshot. An optional checkpoint immediately before the image also serves repeated queries with identical text; it is disabled by default because it adds no reuse on top of divergence-point checkpoints in our traces. The image-token budget independently controls encoding and prefill work. We evaluate its latency--accuracy tradeoff in \secref{sec:eval-image}.
% src: IMAGE_STEP.md sec. 1-3 (M-RoPE positions equal indices before the image; re-query prompt_n 1137); W T-IMG rows
%      "switch unset" 5398 ms vs "=2 + TAIL=0" 4812 ms; results/abl5 (+X on screenshot steps 5355 -> 4945 ms);
%      results/imgbudget.log (510 held-out steps: 5.23 / 3.20 / 2.30 s at 1127 / 512 / 256 tokens, 287 / 280 / 288 right)

\subsection{Idle power release}\label{sec:power}

Keeping the model ready for another task requires retaining its weights in memory. The NPU can enter a low-power state during the wait. \sys tracks operator batches in each session and, after 3~s with no batch submitted or in flight, releases the session's active power requests. It drops the bandwidth vote, powers down HMX, drops the DDR performance mode and restores platform-default clocks and low-power modes. Before submitting the next batch, it reapplies the active votes and waits for the clocks to recover. Power management uses the existing FastRPC power call, independently of the operator queue.

The three-second timer and the app's five-minute unload timer control different resources. The backend timer releases compute and memory-controller votes between bursts. The app timer eventually frees the model's memory after a longer idle period. A request arriving between these deadlines restores the active votes and continues with the loaded weights and cached state.
% src: IDLE_POWER.md sec. 3.3; app unload policy in sec:service.

The backend also reduces wakeups while relaxed. Increasing the queue timeout from 50 to 500~ms reduces periodic wakeups from twenty to two per second; an arriving operator packet wakes the queue immediately. The idle timer accounts for time spent in phone suspend.
% src: IDLE_POWER.md sec. 3.2-3.3 (queue timeout and CLOCK_BOOTTIME_ALARM fallback).
% src: IDLE_POWER.md sec. 3.1-3.4 (mode bit 30 of htp_iface_power; relax: mips_bw cancel, HMX power-down,
%      HAP_power_set_dcvs_v3_init, protected_bus_corners(0), ddr_perf_mode(0); restore replays pwr_req[0..2], 2 ms
%      read-back; 500 ms queue peek; timerfd on CLOCK_BOOTTIME_ALARM)

Power release is synchronized with operator submission and completion. The host waits until all submitted batches have completed and the matrix unit has been relinquished before releasing active votes. Submission and release share a mutex, so a new batch either prevents the release or waits for the NPU to return to its active state. Model weights and inference state remain allocated throughout the transition. Restoring the votes takes 2--4~ms (\secref{sec:eval-energy}).

% src: IDLE_POWER.md sec. 4 (submission sequence, completion visibility, HMX release and mutex ordering).
% src: IDLE_POWER.md sec. 4 (invariants), sec. 5 (exactness), sec. 11 (phone: 61/61 identical answers and top-5
%      log-probs, also after 400 relax/restore cycles; restore RPC mean 2.1-4.2 ms)

% Implementation, Route 1 draft revision 2 (2026-09-30), STYLE_MOBISYS.md sec. 7.
% src: research/route1/paper_data/code/tandem-rt25-vs-b11200.patch (git diff --ignore-cr-at-eol b11200 of the final
%      tree: 24 files, +3,039 / -71; NPU side ggml-hexagon/htp 12 files +2,027, host backend 2 files +748, server +202,
%      model and graph code 8 files +54, tests +8; the plain diff, +11,325 / -8,357, also counts a line-ending conversion
%      of ggml-hexagon.cpp); BASELINE_STACKS.md sec. 8 (NDK r28b, Hexagon SDK 6.6.0.0); test-backend-ops.cpp (6 cases:
%      batches 16 / 33 / 448 x rows 4,100 / 9,216); draftvocab/make_draft_vocab.py; the cache bound is --cache-ram 512
\section{Implementation}\label{sec:impl}

\sys builds on llama.cpp release b11200, adding about 3,000 lines of C and C++ across 24 files. Two thirds of the added code runs on the NPU. The server handles checkpoints and batching; the host backend selects kernels, rewrites fused operators and manages power transitions; the DSP library implements the kernels and power handlers. We compile with Android NDK r28b and Hexagon SDK 6.6.

An offline tool adds the trimmed draft matrix and its original token IDs to the model file. Runtime switches independently control the optimizations. With them disabled, the same build uses the stock serving policy and kernels while retaining the correctness fixes, allowing ablations within one executable.

\noindent\textbf{Service lifecycle.}\label{sec:service} The app loads the server on demand and unloads it after five idle minutes; the prompt cache is bounded at 512\,MiB. Repeated requests exposed a work-queue race in the upstream backend that could execute a task twice, causing hangs or aborts. We fixed the queue synchronization and reported the race to llama.cpp; its maintainers confirmed the issue and merged a fix. Both \sys and our llama.cpp baseline include the correction. A watchdog converts an unresponsive NPU into a restartable server failure.
% src: github.com/ggml-org/llama.cpp PR #29606 (our report and fix, 2026-09-28) and PR #29785 (the maintainers' fix of the
%      root cause, "better fix for the issue described in #29606", merged 2026-10-01, commit 3aa0ce9); LFM2_DET.md for the
%      copy below

\noindent\textbf{DMA coherence.} The convolution writes inputs through DMA, while the state-copy operator reads them through the NPU data cache. Stale cache lines can therefore save a previous layer's state in small batches. The copy path cleans and invalidates the source range through the operating system before reading it.

% src: N sec. 8 (sporadic aborts after a few hundred requests), sec. 11 (hang after ~490 requests; race in the upstream
%      htp/work-queue.c; seqlock re-check; race amplifier: upstream hung on request 1, fixed queue 120/120; llama-bench
%      within 1%); research/npu/patches/README.md (GGML_HEXAGON_WATCHDOG, default 60 s)

% Evaluation, prose revision 2026-10-01. Source notes below record the measurement session for each result.
% Latency savings = 100*(1 - mean(Tandem)/mean(baseline)), calculated before rounding times for display. All
% comparisons: the 2026-10-02 reruns with every mechanism; on the first device with the screen verifiably on
% (exp_screen.sh, labels sc-*; paper_data CSV column screen_off_cap_share = 0). Main v79 text / screenshot: 29.67% /
% 15.94%; v81: 26.76% / 15.94%. Model-control text: 30.94% hybrid, 2.90% full attention; screenshots: 17.04% / 1.75%.
% Sizes 2B / 4B / 9B text: 26.67% / 29.59% / 33.62%. AndroidControl 28.27%; LFM2 against llama.cpp 15.85%; energy 32.9%.
\section{Evaluation}\label{sec:eval}

We measure latency, energy and action fidelity on recorded GUI-agent requests. We vary model size, recurrent architecture, workload and NPU generation to examine how the gains depend on the model and execution platform.

\subsection{Setup}\label{sec:eval-setup}

\noindent\textbf{Devices.} Our primary device is a OnePlus 13T with a Snapdragon 8 Elite, a Hexagon v79 NPU, an Adreno 830 GPU and 16\,GB of memory, running Android 16. A OnePlus 15 provides the next generation: Snapdragon 8 Elite Gen 5, Hexagon v81, Adreno 840 and 16\,GB, also on Android 16. Both processors have eight Oryon CPU cores.

\noindent\textbf{Model and workloads.} The main policy is the fine-tuned Qwen3.5-4B with Q4\_K\_M weights. Text latency uses the 61-request replay trace; screenshot latency uses the 21-screen app set (\secref{sec:char}). A separate \emph{held-out set} contains 510 steps from 54 other episodes in 12 apps, each recorded with its screenshot, so it provides both text and screenshot steps. AndroidControl~\cite{li2024androidcontrol} supplies 300 public test steps from 56 episodes, with English tasks, a different prompt format and the dataset's screenshots. Finally, the agent performs 100 real tasks in 17 apps installed on the phone; the model chooses every action, with at most 20 steps per task.

\noindent\textbf{Baselines.} llama.cpp b11200~\cite{ggml2023llamacpp,ggml2026hexagon} serves the same model on the NPU, GPU and CPU; its NPU backend includes our work-queue fix so it can complete the traces. MNN 3.6.1~\cite{jiang2020mnn} uses its own 4-bit weights and CPU, OpenCL and Vulkan backends. We test speculation where supported and use the fastest NPU configuration for the main comparison. QAIRT 2.45~\cite{qualcomm2026qairt} does not support the recurrent layers in our Qwen3.5-4B model. We separately compare hybrid and full-attention models under llama.cpp and \sys.
Research systems for LLM inference on mobile NPUs target full-attention models (\secref{sec:related}), and MNN's NPU backend runs the linear-attention layers on the CPU, so neither executes our model on the NPU. 
As a research baseline for hybrid prefix reuse, we implement Marconi's checkpoint policy~\cite{pan2025marconi} in the llama.cpp server: it saves the recurrent state where a new prompt branches from the cached sequence and, when the checkpoint list is full, evicts the checkpoint with the lowest combination of recency and computation saved per byte. It uses llama.cpp's kernels and runs without speculation, llama.cpp's fastest configuration.
% src: research/route1/gaps/C1_NPU_SUPPORT.md (survey of 2026-10-17; the MNN and llm.npu statements still need a
%      check of the cited sources before submission). Marconi baseline: research/route1/gaps/marconi/make_marconi.py
%      (LLAMA_MARCONI=1 on llama.cpp b11200 + work-queue fix; only the server library differs from the baseline package).

\noindent\textbf{Measurements.} Step latency covers one model request, excluding app execution, and is averaged over each trace. Requests use greedy JSON decoding without log-probabilities; separate fidelity runs collect first-token log-probabilities. Latency measurements on the primary device use a 15~W thermoelectric cooler; the second device is passively cooled. Paired text configurations alternate, starting fresh servers after the NPU cools below 42\,$^\circ$C; the main NPU comparison averages four runs per engine, and the comparisons with the Marconi policy alternate the three servers. Energy measurements use the phone's fuel gauge on battery without cooling. The screen is held on in every latency and energy measurement except the MNN runs, whose screen state was not recorded, and the image-budget study. Paired runs are compared within each measurement session.

\noindent\textbf{Serving configurations.} The main end-to-end comparisons run \sys in its full configuration, including batch consolidation. Both NPU servers retain eight checkpoints per slot. The ablation measures each configuration after three text warm-up requests, with idle power release disabled. The model comparison uses Q4\_0 weights for both architectures; the size comparison uses the stock, unfine-tuned Qwen3.5 models.
% src: exp_screen.sh main/model2/sizes/arch/ac (2026-10-02 18:15-23:16, first device, lock screen dismissed, "Stay
% awake" on, keep-alive key every 20 s, every run checked against the screen-off CPU caps: all clean) and
% dev2/exp_dev2_tail.sh (second device, "Stay awake"); LLAMA_CKPT_DIVERGE=1 LLAMA_CKPT_TAIL=0, i.e. all mechanisms
% (the server logs show the tail rule active). The morning runs (exp_tail.sh, tl-*) sat at the screen-off CPU caps
% because the lock screen turned the screen off; they are not used. exp_ablation.sh WAKE=2 (three text warm-ups, tail
% consolidation and idle-release switches); results/abl5. MNN rows: exp_mnn.sh 2026-09-30, screen state not recorded.

\subsection{End-to-end latency}\label{sec:eval-e2e}

% F-MAIN: measured server time per step for hybrid-model engines on the same phone; see caption for exceptions.
% Text steps: the 61-request replay trace; screenshot steps: the app set with screenshots (1,127 image tokens).
% src: one session with all of Tandem's mechanisms, without log-probabilities, as the agent requests them
%      (exp_screen.sh main and shots-main, final build rt25, 2026-10-02 18:15-19:47, cooler, screen verifiably on;
%      research/route1/results/screen.log): text steps Tandem 1.370 / 1.378 / 1.377 / 1.382 s (mean 1.377) against
%      llama.cpp NPU without MTP 1.957 / 1.957 / 1.957 / 1.959 s (1.957) in alternating runs; llama.cpp NPU with MTP
%      2.554 s, GPU without MTP 5.066 s, CPU 8 threads without MTP 10.659 s. Screenshot steps (zhlocal.py, which never
%      asks for log-probabilities): Tandem 4.76 / 4.76, llama.cpp NPU 5.65 / 5.67, with MTP 7.47, GPU 23.09, CPU 67.17 s.
%      MNN 3.6.1 (exp_mnn.sh "lat img", 2026-09-30 03:47-05:31, cooler): OpenCL 4.98 s, screenshot 19.14 s with its
%      vision encoder on OpenCL (46.32 s as shipped, encoder on the CPU), Vulkan 6.61 / 53.17 s, CPU 8 threads 11.92 /
%      57.07 s (screenshot rows: 4 steps). llama.cpp's GPU server aborts with MTP (CL_INVALID_WORK_GROUP_SIZE).
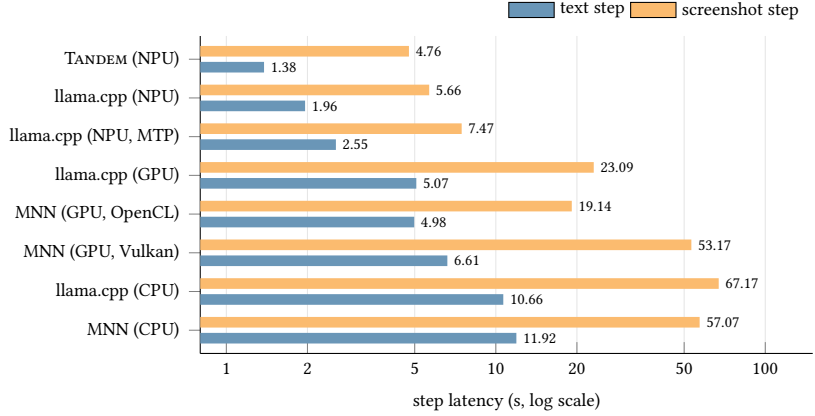
\begin{figure}[t]
\centering
\definecolor{mainText}{HTML}{6a96b7}
\definecolor{mainShot}{HTML}{fbbb6f}
\begin{tikzpicture}
\begin{axis}[
  xbar, scale only axis, width=0.58\columnwidth, height=4.2cm,
  xmode=log, log origin=infty, xmin=0.8, xmax=150,
  xtick={1,2,5,10,20,50,100}, xticklabels={1,2,5,10,20,50,100},
  xlabel={step latency (s, log scale)},
  ytick={1,2,3,4,5,6,7,8}, y dir=reverse, ymin=0.4, ymax=8.6,
  yticklabels={{\sys (NPU)}, {llama.cpp (NPU)}, {llama.cpp (NPU, MTP)}, {llama.cpp (GPU)}, {MNN (GPU, OpenCL)},
    {MNN (GPU, Vulkan)}, {llama.cpp (CPU)}, {MNN (CPU)}},
  tick label style={font=\scriptsize}, label style={font=\scriptsize},
  bar width=4pt, xmajorgrids, grid style={black!10}, axis y line*=left, axis x line*=bottom,
  point meta=rawx, nodes near coords={\pgfmathprintnumber[fixed,precision=2]{\pgfplotspointmeta}},
  every node near coord/.append style={font=\tiny, anchor=west, xshift=-1pt},
  legend style={font=\scriptsize, draw=none, fill=none, at={(1,1.02)}, anchor=south east, legend columns=2},
  legend cell align=left, area legend,
]
\addplot[fill=mainText, draw=none] coordinates {(1.38,1) (1.96,2) (2.55,3) (5.07,4) (4.98,5) (6.61,6) (10.66,7) (11.92,8)};
\addplot[fill=mainShot, draw=none] coordinates {(4.76,1) (5.66,2) (7.47,3) (23.09,4) (19.14,5) (53.17,6) (67.17,7) (57.07,8)};
\legend{text step, screenshot step}
\end{axis}
\end{tikzpicture}
\caption{Step latency on a Snapdragon 8 Elite with the fine-tuned Qwen3.5-4B in Q4\_K\_M: text uses the 61-request replay trace; screenshots use the 21-screen app set. MNN screenshot bars average 4 of the 21 steps.}
\Description{Horizontal bar chart on a log scale. Tandem on the NPU has the shortest bars for the hybrid model, 1.38 seconds per text step and 4.76 seconds per screenshot step, followed by llama.cpp on the NPU with 1.96 and 5.66 seconds, and with speculation 2.55 and 7.47 seconds; the GPU and CPU engines take 5 to 12 seconds per text step and 19 to 67 seconds per screenshot step.}
\label{fig:main}
\end{figure}

\sys reduces text-step latency by 30\%, from 1.96 to 1.38~s, relative to llama.cpp's fastest NPU configuration (Figure~\ref{fig:main}). Screenshot latency falls by 16\%, from 5.66 to 4.76~s. The smaller screenshot speedup reflects the encoder, which both runtimes execute on the NPU and which accounts for about a third of the step.

\noindent\textbf{NPU execution.} llama.cpp is fastest with speculation disabled. Enabling it raises text-step latency to 2.55~s because recurrent-state snapshots slow prefill and the verification kernel scales poorly with the draft count. \sys reduces both costs, allowing speculative execution to shorten action generation alongside the saving from context reuse.

\noindent\textbf{GPU and CPU execution.} The fastest measured GPU configuration, MNN OpenCL, takes 4.98~s per text step and 19.1~s per screenshot step with its encoder on the GPU. \sys reduces these times by 3.6$\times$ and 4.0$\times$. llama.cpp's GPU backend has similar text latency but longer screenshot latency. Both engines take 11--12~s per text step on the CPU. These configurations lack the NPU execution path for the model's recurrent layers.

% src: figures/comparison.tex (exp_screen.sh main and shots-main of 2026-10-02 18:15-19:47, final build rt25 with all
%      mechanisms, cooler, screen verifiably on, without log-probabilities, research/route1/results/screen.log: Tandem
%      1.370 / 1.378 / 1.377 / 1.382 s, llama.cpp NPU 1.957 / 1.957 / 1.957 / 1.959 s, MTP 2.554, GPU 5.066, CPU 10.659
%      s; screenshots 4.76 / 4.76 against 5.65 / 5.67 s, MTP 7.47, GPU 23.09, CPU 67.17 s; MNN rows exp_mnn.sh of
%      2026-09-30); Tandem's encoder share: IMAGE_STEP.md sec. 5 (1.7 s of about 5 s)

\begin{table*}[t]
\centering
% \begin{updblock}
\caption{Server time per step (s) on the main and larger workloads. Each block of rows was measured in one session with alternating runs. Marconi denotes llama.cpp with Marconi's checkpoint policy. Gain is \sys's latency reduction relative to llama.cpp.}
\label{tab:workloads}
\small
\begin{tabular}{lccccc}
\toprule
Workload & Steps & llama.cpp & Marconi & \sys & Gain \\
\midrule
Replay trace, text & 61 & 2.01 & 1.82 & 1.44 & 28\% \\
App set, screenshots & 21 & 5.76 & 5.25 & 4.87 & 15\% \\
Held-out set, text & 510 & 2.02 & 1.86 & 1.45 & 28\% \\
\midrule
AndroidControl, text & 300 & 1.61 & 1.51 & 1.16 & 28\% \\
AndroidControl, screenshots & 300 & 3.95 & 3.86 & 3.44 & 13\% \\
\midrule
Held-out set, screenshots & 510 & 6.32 & -- & 5.35 & 15\% \\
\bottomrule
\end{tabular}
% \end{updblock}
\end{table*}
% src: research/route1/results/gaps/paper_numbers.txt (research/route1/gaps/paper_numbers.py) and summary.md; final build
%      rt25 with all mechanisms, cooler, screen verifiably on, without log-probabilities. Rows 1-3: one session on
%      2026-10-17/18 (gaps/exp_mc.sh, order T U M M U T): replay 2.010 / 1.816 / 1.439 s (prompt 541 / 477 / 464 tok),
%      app set 5.758 / 5.246 / 4.872 s (1778 / 1512 / 1512 tok, 20/21 right under all three), held-out text 2.020 /
%      1.860 / 1.449 s (475 / 439 / 441 tok; 265 / 267 / 265 of 510 right). Held-out screenshots: exp_gaps.sh held-img
%      (2026-10-17, T U U T): 6.320 vs 5.346 s (T runs 5.248 / 5.445; the second ran after the phone was repositioned),
%      288 / 287 of 510 right. Rows 4-5: one session on 2026-10-19 08:50-11:43 (gaps/block9.sh, exp_mc.sh mc-actext
%      mc-acimg, order T U M M U T): AndroidControl text 1.610 / 1.508 / 1.163 s (393 / 362 / 366 tok, step success
%      182/300 under all three), screenshots 3.951 / 3.864 / 3.440 s (1121 / 1084 / 1089 tok, 184/300 under all three).
%      Earlier two-engine sessions: AndroidControl text 2026-10-02 (sc-ac-*, 1.570 vs 1.126 s), screenshots
%      exp_gaps.sh ac-img 2026-10-17 (4.047 vs 3.520 s, 1121 vs 1089 tok, type 238/300 and step success 184/300 under
%      both). The first held-out text session (2026-10-17): 1.969 vs 1.448 s (26.4%).

% \begin{updblock}
\noindent\textbf{Larger workloads.} The gains hold beyond the replay trace (Table~\ref{tab:workloads}). On the 510 held-out steps, \sys reduces text-step latency by 26--28\% in two sessions and screenshot latency by 15\%, with the accuracy of llama.cpp: 265 of 510 text steps and 287 versus 288 screenshot steps are correct. The reduction shrinks as new prompt work grows, from 32\% for steps that require fewer than 300 new tokens under llama.cpp to 22\% for steps that require more than 700, because tokens that neither runtime can reuse dominate those steps. On AndroidControl, the separate system message already gives llama.cpp a checkpoint at the user-message boundary, so \sys processes only 3--7\% fewer prompt tokens; it still reduces latency by 28\% on text steps and 13\% on screenshot steps, mostly through faster decoding.
% \end{updblock}
% src: held-out text buckets by llama.cpp's prompt tokens (paper_numbers.txt): <300 149 steps 1.391 vs 0.946 s (32.0%),
%      300-400 31.3%, 400-500 27.5%, 500-700 23.0%, >700 97 steps 2.883 vs 2.244 s (22.2%). AndroidControl prompt tokens:
%      text 366 vs 393 (7%), screenshots 1089 vs 1121 (3%).

\subsection{Comparison with hybrid prefix caching}\label{sec:eval-marconi}

% \begin{updblock}
Marconi's checkpoint policy recovers the prefix reuse but not the decoding cost (Table~\ref{tab:workloads}). On the replay trace it processes 477 new prompt tokens per step, close to \sys's 464 and below llama.cpp's 541, and lowers text-step latency by 10\% relative to llama.cpp. On the held-out text steps it processes slightly fewer tokens than \sys, 439 versus 441, because it checkpoints every branch point while \sys requires 32 tokens of new context. \sys nevertheless remains 21--23\% faster on text steps and 7--11\% faster on screenshot steps. The difference is decoding: \sys generates a token in 44~ms, llama.cpp and Marconi in 66~ms. A checkpoint policy alone cannot make speculation profitable on a hybrid model; that also requires rollback-aware prefill, the shared verification kernel and the trimmed draft head (\secref{sec:eval-ablation}).
% \end{updblock}
% src: as tab:workloads rows 1-5 (Tandem vs Marconi: replay 20.7%, app set 7.1%, held-out text 22.1%, AndroidControl
%      text 22.9%, screenshots 11.0%); decode 43.7-46.4 ms per token for Tandem, 65.4-66.9 for llama.cpp and Marconi
%      (rows 1-3; the AndroidControl client does not record decode timings).

\subsection{Where the gains come from}\label{sec:eval-ablation}

% Figure 6: cumulative single-build ablation, text steps of the replay trace.
% src: research/route1/results/abl5/summary.md (exp_ablation.sh WAKE=2, rt25, cooler, block 1, 2026-10-02 23:16 -
%      2026-10-03 01:29, without log-probabilities, screen verifiably on in every run), cumulative rows V, PV, PCV, PCGV,
%      PCGWV, PCGWMV, PCGWMHV, PCGWMHRV, PCGWMHRDV, PCGWMHRDTV, PCGWMHRDTXV (text step 1977, 1957, 1920, 1892, 1850, 2516,
%      1987, 1770, 1647, 1523, 1408 ms; screenshot 5915, 5846, 5733, 5726, 5575, 7497, 5850, 5605, 5506, 5355, 4945 ms).
%      Left out: the prompt split S (changes the model's input) and the checkpoint before the image I (off in the final
%      configuration; +8 ms text, +10 ms screenshot on top of PCGWMHRDTXV). Earlier runs: results/abl4 (2026-10-01,
%      screen woken but not kept on) and results/abl3 (2026-09-30, timed with log-probabilities).
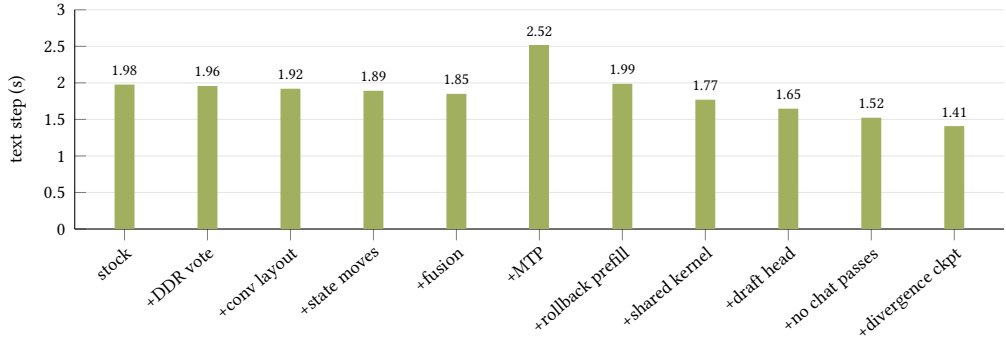
\begin{figure}[t]
\centering
\definecolor{ablBar}{HTML}{a1b05e}
\definecolor{ablSpike}{HTML}{a1b05e}
\begin{tikzpicture}
\begin{axis}[
  ybar, scale only axis, width=0.88\columnwidth, height=2.9cm,
  ymin=0, ymax=3.0, ytick={0,0.5,1,1.5,2,2.5,3},
  ylabel={text step (s)}, xtick={1,...,11}, xmin=0.4, xmax=11.6,
  xticklabels={stock, +DDR vote, +conv layout, +state moves, +fusion, +MTP, +rollback prefill, +shared kernel,
    +draft head, +no chat passes, +divergence ckpt},
  xticklabel style={rotate=40, anchor=north east, font=\scriptsize, inner sep=1pt},
  tick label style={font=\scriptsize}, label style={font=\scriptsize},
  bar width=7.5pt, bar shift=0pt, ymajorgrids, grid style={black!10}, axis x line*=bottom, axis y line*=left,
  point meta=rawy, nodes near coords={\pgfmathprintnumber[fixed,precision=2]{\pgfplotspointmeta}},
  every node near coord/.append style={font=\tiny},
]
\addplot[fill=ablBar, draw=none] coordinates {(1,1.977) (2,1.957) (3,1.920) (4,1.892) (5,1.850) (7,1.987) (8,1.770) (9,1.647) (10,1.523) (11,1.408)};
\addplot[fill=ablSpike, draw=none, bar shift=0pt] coordinates {(6,2.516)};
\end{axis}
\end{tikzpicture}
\caption{Cumulative ablation on the replay trace after three text warm-up requests, using runtime switches in one executable. Idle power release is disabled. The last two additions enable batch consolidation and divergence-point checkpoints.}
\Description{Bar chart of the text step time as mechanisms are added: 1.98 seconds for the stock engine, falling to 1.85 seconds with the power vote, the data-movement mechanisms and the fused operator, rising to 2.52 seconds when speculation is switched on, and falling to 1.41 seconds with the rollback-aware prefill, the column-shared kernel, the trimmed draft head, the removal of chat-only passes and divergence-point checkpoints.}
\label{fig:ablation}
\end{figure}

Speculation becomes beneficial only after reducing both rollback and decode costs (Figure~\ref{fig:ablation}). This experiment uses the warmed replay trace and runtime switches in one executable (\secref{sec:eval-setup}). Starting with the stock policy and kernels, prefill and memory-traffic optimizations reduce text-step latency from 1.98 to 1.85~s. Adding MTP increases it by 0.67~s. Rollback-aware prefill recovers most of that loss; column-shared verification and the trimmed draft head reduce the step to 1.65~s. Batch consolidation and divergence-point checkpoints bring it to 1.41~s.

Removing one mechanism at a time measures its contribution within the full configuration. Rollback-aware prefill has the largest effect: 0.46~s per text step and 1.42~s per screenshot step. The shared verification kernel and the trimmed draft head save 0.18 and 0.14~s per text step, batch consolidation and divergence-point checkpoints another 0.17 and 0.08~s. Batching policies also change kernel work, so these individual reductions need not sum to the total latency saving.

For screenshots, placing a checkpoint at the divergence point restores 266 text tokens and saves 0.41~s. A checkpoint immediately before the image restores the same prefix on this workload, adding little when divergence-point checkpoints are already enabled. We therefore leave it off.
% src: research/route1/results/abl5/summary.md (exp_ablation.sh WAKE=2, rt25, cooler, block 1, 2026-10-02 23:16 -
%      2026-10-03 01:29, without log-probabilities, screen verifiably on in all 23 runs): cumulative rows as in Figure 6;
%      leave-one-out text -H +460 [+385, +537], -M +183, -R +176, -T +170, -D +136, -X +79, -C +41, -W +18, -G +14,
%      -P 0, -I -8 ms; screenshot V 5915 -> final 4945 ms, -V +50.4 s, -H +1422 ms; screenshot prompt tokens 1778
%      without X and I, 1512 with X, with I or with both (+X -410 ms, +I after X +10 ms, -X with I on -9 ms); final
%      configuration without I (PCGWMHRDTXV): text 1408 ms, screenshot 4945 ms. The 2026-10-01 run (results/abl4) woke
%      the screen but did not keep it on (14 of 23 runs started with it dozing or asleep; ColorOS caps the CPU with the
%      screen off); D and P saved more there (-221 and -24 ms).

\subsection{Hybrid versus full attention}\label{sec:eval-model}

\begin{table}[t]
\centering
\caption{Server time per step (s) of the hybrid policy and of the full-attention Qwen3-VL-4B, both with Q4\_0 weights, on llama.cpp (fastest configuration) and on \sys, each step type measured in one session. Text steps: the replay trace; screenshot steps: the app set, at each model's native resolution. Gain denotes the relative latency reduction.}
\label{tab:model}
\small
\begin{tabular}{llccc}
\toprule
Step & Model & llama.cpp & \sys & Gain \\
\midrule
Text & full attention & 1.33 & 1.29 & 3\% \\
 & hybrid & 1.90 & 1.31 & 31\% \\
 & hybrid / attention & 1.43$\times$ & 1.01$\times$ & \\
\midrule
Screenshot & full attention & 5.09 & 5.00 & 2\% \\
 & hybrid & 5.56 & 4.61 & 17\% \\
 & hybrid / attention & 1.09$\times$ & 0.92$\times$ & \\
\bottomrule
\end{tabular}
\end{table}
% src: text steps: research/route1/results/screen.log (exp_screen.sh model2, final build rt25 with all mechanisms,
%      2026-10-02 20:22-20:41, without log-probabilities, screen verifiably on, order HT HU FT FU FU FT HU HT): hybrid
%      Tandem 1.298 / 1.327, hybrid llama.cpp 1.906 / 1.895, full attention Tandem 1.297 / 1.292, full attention
%      llama.cpp 1.340 / 1.326 s; prompt_n 463.9 / 541.1 / 424.6 / 424.6; decode 40.2 / 63.6 / 55.6 / 56.8 ms per token.
%      Screenshot steps (zhlocal.py, no log-probabilities): exp_screen.sh shots-model2, 20:41-21:03: 4.61 / 5.56 / 5.00 /
%      5.09 s (19/21 right for the hybrid and 12/21 for the full-attention model under both engines). Hybrid / full
%      attention from unrounded means: llama.cpp 1.9005 / 1.3330 = 1.426 (43%), Tandem 1.3125 / 1.2944 = 1.014 (1.4%).

With both models quantized to Q4\_0, \sys reduces text-step latency by 31\% for the hybrid model and 3\% for full attention (Table~\ref{tab:model}). The hybrid's 43\% latency penalty under llama.cpp falls to 1.4\%.

Checkpoint placement reduces the hybrid's new prompt work from 541 to 464 tokens per step, compared with 425 for the full-attention model. MTP then lowers its decode time to 40~ms per token. Full attention already truncates its KV cache at the shared prefix and has no MTP head in this comparison; its gain comes from the shared backend changes.

\subsection{Screenshot steps}\label{sec:eval-image}

\begin{table}[t]
\centering
\caption{Image budget on the 510-step held-out set, with \sys in its full configuration}. The text-only row uses the same requests without screenshots.
\label{tab:image-budget}
\small
\begin{tabular}{lccc}
\toprule
Image tokens & Step (s) & Correct & Accuracy \\
\midrule
Text only & 1.45 & 265 & 52.0\% \\
1,127 (default) & 5.35 & 287 & 56.3\% \\
512 & 3.21 & 280 & 54.9\% \\
256 & 2.45 & 288 & 56.5\% \\
\bottomrule
\end{tabular}
\end{table}
% src: final build (2026-10-17): text-only = held-out text runs gp-ht-T (1.449 / 1.447 s, 265), 1127 = gp-hi-T
%      (5.248 / 5.445 s, 287), 512 = gp-ib512-T (3.208 s, 280), 256 = gp-ib256-T (2.448 s, 288); paper_numbers.txt.
%      The rt22 version (2026-09-30, below) gave 1.41 / 5.23 / 3.20 / 2.30 s and 267 / 287 / 280 / 288.

On the 510-step held-out set, reducing the image budget from 1,127 to 256 tokens lowers latency from 5.35 to 2.45~s, a 2.2$\times$ speedup (Table~\ref{tab:image-budget}). Accuracy is 56.3\% at the default budget and 56.5\% at 256 tokens; McNemar's test detects no significant pairwise differences among the three image budgets ($p\ge0.17$). Text-only input reaches 52.0\%; the default and 256-token settings improve on it ($p=0.015$ and $0.006$).
% src: final build, exact McNemar (paper_numbers.txt): 1127 vs 512 16/9 p=0.230, 1127 vs 256 19/20 p=1.0, 512 vs 256
%      9/17 p=0.169; image vs text 49/27 p=0.015 (1127), 40/25 p=0.082 (512), 44/21 p=0.006 (256).
% src (rt22 version): research/route1/results/imgbudget.log and ib-img{1127,512,256}.jsonl (exp_imgbudget.sh, 2026-09-30, rt22 with the
%      full configuration of that date, cooler; 510 held-out steps of local_v3 test with screenshots): 287 / 280 / 288 of
%      510, 5.23 / 3.20 / 2.30 s per step, 1600 / 958 / 693 prompt tokens; text-only 267 of 510 (52.4%), 1.41 s
%      (androidcontrol/results/ib-text-local.jsonl). Exact McNemar: 1127 vs 512 17/10 p=0.25, 1127 vs 256 20/21 p=1.0,
%      512 vs 256 10/18 p=0.18; image vs text 47/27 p=0.027 (1127), 36/23 p=0.117 (512), 42/21 p=0.011 (256)
% src: results/abl5 (+X on screenshot steps: 1778 -> 1512 prompt tokens, 5355 -> 4945 ms); W T-B1 (5.49 / 3.39 / 2.38 s;
%      20 / 19 / 19 of 21)

\subsection{Sensitivity to configuration}\label{sec:eval-sensitivity}

% F-SENS: one setting varied at a time on the 61-request replay trace, final build with all mechanisms, cooler, screen
% verifiably on, without log-probabilities (research/route1/gaps/exp_gaps.sh sweep-ckpt, sweep-draft, draft-rep,
% sweep-cache, idle; 2026-10-17 14:05-19:06). Values: research/route1/results/gaps/paper_numbers.txt.
%   (a) checkpoints per slot: Tandem 1.650 1.513 1.469 1.435 1.440 s; llama.cpp 2.295 2.166 2.027 1.997 1.993 s
%   (b) draft length (Tandem): 1.540 1.371 1.434 1.400 1.537 s (lengths 2, 3, 4: means of 3, 2, 3 runs);
%       draft-token acceptance 0.88 0.82 0.72 0.65 0.50
%   (c) prompt-cache cap 0 128 256 512 1024 4096 MiB: Tandem 1.378 1.400 1.433 1.438 1.466 1.466 s; llama.cpp 1.955
%       1.968 1.962 1.997 1.993 2.013 s
%   (d) NPU power-collapsed share of the replay (debugfs) for pauses of 2, 5, 10 s after each response: threshold 1 s
%       38 67 82 %, 3 s (default) 0 36 64 %, 10 s 0 0 3 %; without release 0 % throughout
\begin{figure}[t]
% \begin{updblock}
\centering
\definecolor{sensT}{HTML}{2A78D6}
\definecolor{sensU}{HTML}{EB6834}
\pgfplotsset{sens/.style={scale only axis, width=0.36\columnwidth, height=2.6cm,
  tick label style={font=\scriptsize}, label style={font=\scriptsize}, title style={font=\scriptsize},
  ymajorgrids, grid style={black!10}, axis x line*=bottom, axis y line*=left,
  legend style={font=\tiny, draw=none, fill=none}, legend cell align=left, every axis plot/.append style={thick}}}
\begin{tikzpicture}
\begin{axis}[sens, title={(a) checkpoints per slot}, xtick={1,2,3,4,5}, xticklabels={1,2,4,8,16},
  ylabel={s per step}, ymin=1.2, ymax=2.4, legend style={at={(1,1)}, anchor=north east}]
\addplot[sensT, mark=*, mark size=1.2pt] coordinates {(1,1.650) (2,1.513) (3,1.469) (4,1.435) (5,1.440)};
\addplot[sensU, mark=square*, mark size=1.2pt] coordinates {(1,2.295) (2,2.166) (3,2.027) (4,1.997) (5,1.993)};
\legend{\sys, llama.cpp}
\end{axis}
\end{tikzpicture}\hfill
\begin{tikzpicture}
\begin{axis}[sens, title={(b) draft length (\sys)}, xtick={1,2,3,4,6}, ymin=1.3, ymax=1.6, ylabel={s per step},
  nodes near coords, point meta=explicit symbolic, every node near coord/.append style={font=\tiny, anchor=south}]
\addplot[sensT, mark=*, mark size=1.2pt] coordinates {(1,1.540) [0.88] (2,1.371) [0.82] (3,1.434) [0.72] (4,1.400) [0.65] (6,1.537) [0.50]};
\end{axis}
\end{tikzpicture}

\vspace{2pt}
\begin{tikzpicture}
\begin{axis}[sens, title={(c) prompt-cache cap (MiB)}, xtick={1,2,3,4,5,6}, xticklabels={0,128,256,512,1k,4k},
  ylabel={s per step}, ymin=1.2, ymax=2.4]
\addplot[sensT, mark=*, mark size=1.2pt] coordinates {(1,1.378) (2,1.400) (3,1.433) (4,1.438) (5,1.466) (6,1.466)};
\addplot[sensU, mark=square*, mark size=1.2pt] coordinates {(1,1.955) (2,1.968) (3,1.962) (4,1.997) (5,1.993) (6,2.013)};
\end{axis}
\end{tikzpicture}\hfill
\begin{tikzpicture}
\begin{axis}[sens, title={(d) NPU asleep during pauses}, xtick={2,5,10}, xlabel={pause between requests (s)},
  ylabel={power-collapsed (\%)}, ymin=0, ymax=100, legend style={at={(0,1)}, anchor=north west}]
\addplot[black!70, mark=triangle*, mark size=1.4pt] coordinates {(2,38) (5,67) (10,82)};
\addplot[sensT, mark=*, mark size=1.2pt] coordinates {(2,0) (5,36) (10,64)};
\addplot[black!35, mark=diamond*, mark size=1.4pt] coordinates {(2,0) (5,0) (10,3)};
\legend{1 s threshold, 3 s (default), 10 s}
\end{axis}
\end{tikzpicture}
\caption{Sensitivity on the replay trace, one setting varied at a time. (a)--(c) server time per step; the labels in (b) give the fraction of draft tokens accepted. (d) Share of the replay in which the NPU is power-collapsed when the agent pauses after each response, for three idle-release thresholds; without release it is 0.}
\Description{Four small line plots. (a) Step time falls from 1.65 to 1.44 seconds for Tandem and from 2.30 to 1.99 seconds for llama.cpp as checkpoints per slot grow from 1 to 8, with no change at 16. (b) Tandem's step time is lowest at draft length 2, 1.37 seconds, and highest at lengths 1 and 6, about 1.54 seconds; acceptance falls from 0.88 to 0.50. (c) Step time varies little with the prompt-cache cap for both engines. (d) The NPU sleeps for 0 to 82 percent of the replay depending on the pause and the threshold.}
\label{fig:sensitivity}
% \end{updblock}
\end{figure}
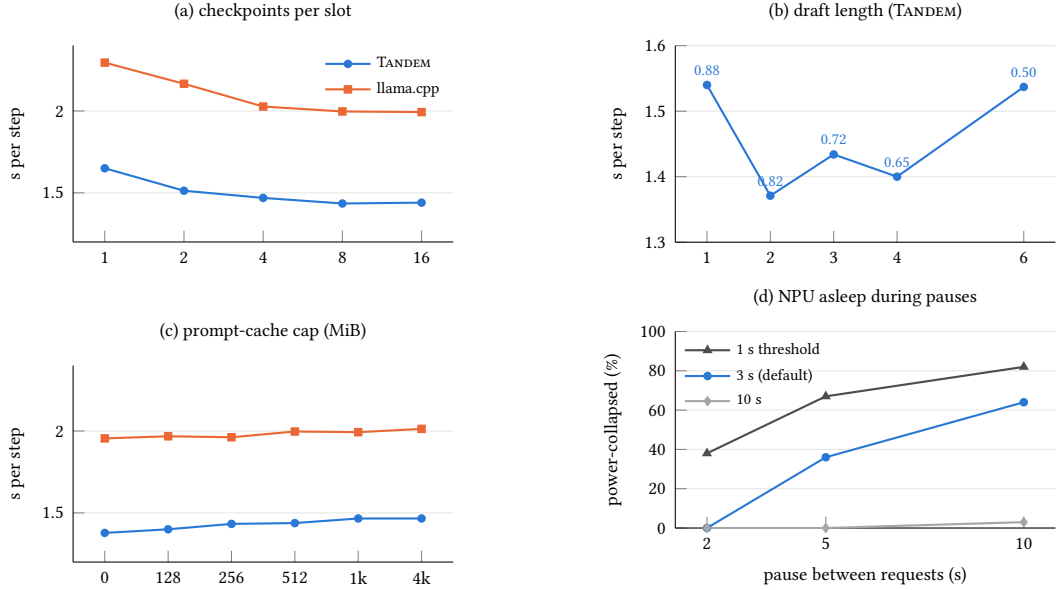

% \begin{updblock}
Figure~\ref{fig:sensitivity} varies one setting at a time on the replay trace. Eight checkpoints per slot suffice: more checkpoints change neither prompt work nor latency, and with a single checkpoint per slot \sys still runs 28\% faster than llama.cpp with the same limit. The default draft length of three tokens is within 5\% of the best length, two: longer drafts are accepted less often, 82\% of draft tokens at two and 50\% at six, while single-token drafts save fewer decoding passes. The prompt-cache cap changes prompt work for neither engine, because one agent's sequential requests reuse the checkpoints in their own slot; latency varies by less than 7\% across caps from 0 to 4~GiB. Idle power release trades a short wake-up for NPU sleep. With the default 3-second threshold, pauses of 5 and 10~s between requests keep the NPU power-collapsed for 36\% and 64\% of the replay at less than 0.5\% added latency; a 1-second threshold raises these shares to 67\% and 82\% at 1.3--1.6\% added latency.
% \end{updblock}
% src: research/route1/results/gaps/paper_numbers.txt (exp_gaps.sh sweep-ckpt, sweep-draft, sweep-cache, idle and
%      draft-rep, 2026-10-17 14:05-19:06, one session, cooler, screen verifiably on). Checkpoints 1/2/4/8/16: Tandem 1.650 /
%      1.513 / 1.469 / 1.435 / 1.440 s, llama.cpp 2.295 / 2.166 / 2.027 / 1.997 / 1.993 s. Draft n_max 1/2/3/4/6: 1.540 /
%      1.371 / 1.434 / 1.400 / 1.537 s (2, 3, 4: means of 3, 2, 3 runs; one n_max 3 run excluded because a video app was
%      in the foreground), acceptance 0.88 / 0.82 / 0.72 / 0.65 / 0.50; 3 vs 2 +63 ms per request [36, 87]. Cache cap
%      0-4096 MiB: Tandem 1.378-1.466 s (464-472 tok), llama.cpp 1.955-2.013 s (541 tok). Idle (pause, threshold): NPU
%      power-collapsed share from debugfs master_cdsp_stats over the replay; latency against the threshold-0 run of the
%      same pause: 5 s / 3 s +6 ms, 10 s / 3 s +2 ms, 5 s / 1 s +22 ms, 10 s / 1 s +18 ms.

\subsection{Energy and idle power}\label{sec:eval-energy}

\begin{table}[t]
\centering
\caption{Server time, energy above idle power and the phone's mean power while serving, per text step of the app set, on battery without any cooling, for the engines that run the hybrid model on the NPU, GPU and CPU; means over the runs of one session. The GPU rows come from a separate block, in which the \sys reference run used 11.2~J.}
\label{tab:energy}
\small
\begin{tabular}{lccc}
\toprule
Engine & Step (s) & Energy (J) & Power (W) \\
\midrule
\sys (NPU) & 1.24 & 10.1 & 9.0 \\
llama.cpp (NPU) & 1.91 & 15.1 & 9.5 \\
llama.cpp (NPU, MTP) & 2.40 & 17.1 & 8.9 \\
MNN (GPU, OpenCL) & 5.14 & 35.6 & 8.1 \\
llama.cpp (GPU) & 5.21 & 33.0 & 7.3 \\
llama.cpp (CPU) & 21.2 & 86.2 & 5.1 \\
\bottomrule
\end{tabular}
\end{table}
% src: research/route1/paper_data/energy.csv, rows with tandem_tail_rule = on (exp_energy_tail.sh, 2026-10-02
%      13:41-15:22, final build rt25 with all mechanisms, USB unplugged, no cooler, screen on; tools/analyze_energy.py,
%      energy above the first 2-27 s of each segment). Blocks a (T U M C, M U T) and b (T U U T): Tandem 9.07 / 10.62 /
%      10.82 / 9.88 J (1.235 s, 9.05 W), llama.cpp NPU without MTP 14.77 / 15.31 / 15.27 / 14.87 J (1.908 s, 9.50 W) ->
%      32.9% less (block a 34.5%, block b 31.3%); with MTP 16.49 / 17.78 J (2.402 s, 8.86 W); CPU 21.19 s, 86.24 J,
%      5.06 W. GPU block (Tandem reference 1.24 s, 11.19 J, run last on the warmest phone; counting it too gives 31.5%):
%      MNN 3.6.1 OpenCL 5.14 s, 35.56 J, 8.07 W (63 requests, windows rebuilt from its wall times), llama.cpp OpenCL
%      without MTP 5.21 s, 33.03 J, 7.28 W. MTP vs no MTP (both block a): +26% time, +14% energy. Against the Tandem
%      run of the same block: MNN 35.56 / 11.19 = 3.18x, llama.cpp GPU 33.03 / 11.19 = 2.95x; CPU (block a) 86.24 /
%      9.845 (mean of block a's two Tandem runs) = 8.76x. The sessions of 2026-09-30 and 10-01 without batch consolidation gave Tandem
%      11.1 J against 15.0 J (26%): llama.cpp is unchanged, Tandem's prefill per app-set step fell from 619-664 to
%      489-530 ms for the same 344 tokens.

We measure energy on the app set, on battery without cooling (Table~\ref{tab:energy}). \sys uses 10.1~J above idle per text step, 33\% less than the fastest llama.cpp NPU configuration. Both draw roughly 9~W while active, so the saving follows the reduction in execution time. GPU and CPU backends draw less power but run longer. Against the \sys run in the same measurement block, the GPU backends consume 3.0--3.2$\times$ as much energy and the CPU backend 8.8$\times$. Enabling speculation in llama.cpp raises energy by 14\%, consistent with its longer execution.

\noindent\textbf{Sustained load.} During ten minutes of back-to-back requests without cooling, thermal throttling increases mean step latency from 1.18~s in the first minute to 1.43~s in the last. Energy averages 10.6~J per step.
% src: research/route1/results/energy-tail-a/es3-sus-sustain (rt25 with all mechanisms, 441 steps in 600.2 s, mean
%      1.36 s; first 60 s 1.18 s, last 60 s 1.43 s; 10.59 J net per step, 8.70 W; nsp up to 106.5 C; skin 40.1 ->
%      49.6 C; nsp 49.8 C at t+710 s); without batch consolidation (2026-09-30): 1.37 -> 1.58 s, 11.57 J

\noindent=\textbf{Real tasks.} On the 100 real tasks, run on battery with each engine, \sys lowers model time per step by 28\%, from 1.75 to 1.25~s, and the phone's total energy per step by 20\%, from 25.2 to 20.1~J. The total includes the screen, the app and the agent's observation of the phone; its saving of 5.1~J per step matches the server-level saving on the app set (Table~\ref{tab:energy}).
% src: research/route1/gaps/e2e_analyze.py on results/gaps/e2e/gp-e2e-q*-{T,U} (gaps/exp_e2e.sh s1 + s2, 2026-10-18
%      03:00-04:34 and 05:15-07:01, final build with all mechanisms vs llama.cpp NPU without speculation, charger
%      unplugged, no cooler, screen on; tasks research/teacher/tasks_e2e100.jsonl, quarters of 25 in ABBA order; energy
%      = battery current x voltage from the fuel gauge every 0.2 s, integrated over each task, divided by its steps):
%      model time per step 1.249 vs 1.747 s, ratio 0.715 (95% CI 0.680-0.754 over tasks); energy per step 20.06 vs
%      25.20 J, ratio 0.796 (95% CI 0.735-0.861); 1,327 vs 1,222 steps. Server-level saving in tab:energy:
%      15.1 - 10.1 = 5.0 J per step. Task success is not reported (a property of the model, not of the runtime).

\noindent\textbf{Idle power release.} Releasing session votes lowers screen-on idle power from 0.89 to 0.42~W and screen-off standby power from 0.76 to 0.22~W, close to the no-server measurements of 0.48 and 0.25~W. Releasing only the sleep-disable vote recovers about half the saving because clock, bus and DRAM votes also keep the power domain active. Releasing all votes allows the NPU to remain power-collapsed throughout the idle measurement window.

A request following a 30-second pause takes 21~ms longer overall, about 1.5\% of a step, including 2--4~ms to restore active votes. A six-second pause produces no measurable latency increase. Answers and first-token log-probabilities remain bit-identical, including after 400 consecutive release cycles.

% src: illustrative budget from tab:energy, 10*(15.05-10.10)=49.5 J; IDLE_POWER.md sec. 11.1,
%      (0.76-0.22)*(300-3)=160.38 J. This is a calculation from measured rates, not a new task measurement.
% src: W sec. 5.2 (E3 10.5 / 11.4 J per step); IDLE_POWER.md sec. 11.1 (battery: screen on 0.48 / 0.89 / 0.42 W, screen off 0.25 / 0.76 / 0.22 W, sleep vote only 0.68 W); sec. 11
%      (debugfs master_cdsp_stats power-collapse time over 60 s: none +60, rt22 +0, rt23 +61; restore RPC mean 2.1-4.2 ms;
%      first request after idle: 6 s -0.4 ms [-12.3, +10.4], 30 s +21.0 ms [+8.8, +33.0]; 640 -> 656 tok/s; bit identity
%      in the default, off, stress and gap runs); 21.0 ms of the 1.377 s text step (fig:main) = 1.5%

% \subsection{Model sizes and workloads}\label{sec:eval-sizes}
\subsection{Generality across models, workloads, and devices}
\subsubsection{Model sizes}\label{sec:eval-sizes}

\begin{table}[t]
\centering
\caption{Server time per step (s) of the stock Qwen3.5 models on the NPU, with the cooler: text steps of the replay trace and screenshot steps of the app set. llama.cpp runs in its fastest configuration, without speculation. Gain denotes the relative text-step latency reduction.}
\label{tab:sizes}
\footnotesize
\setlength{\tabcolsep}{4.5pt}
\begin{tabular}{lccccc}
\toprule
 & \multicolumn{3}{c}{Text step} & \multicolumn{2}{c}{Screenshot step} \\
\cmidrule(lr){2-4}\cmidrule(lr){5-6}
Model & llama.cpp & \sys & Gain & llama.cpp & \sys \\
\midrule
2B & 0.99 & 0.73 & 27\% & 3.46 & 3.11 \\
4B & 1.79 & 1.26 & 30\% & 5.94 & 4.84 \\
9B & 2.99 & 1.99 & 34\% & 20.3 & 18.1 \\
\bottomrule
\end{tabular}
\end{table}
% src: text steps: research/route1/results/screen.log (exp_screen.sh sizes, final build rt25 with all mechanisms,
%      2026-10-02 21:03-21:43, without log-probabilities, screen verifiably on, order T U U T M per size; stock unsloth
%      MTP GGUFs with the same 32,768-token draft vocabulary as the fine-tuned model, the 9B draft head cut from its
%      untied output matrix): Tandem 0.725 / 0.729, 1.262 / 1.258, 1.967 / 2.007 s; llama.cpp without MTP 0.989 /
%      0.993, 1.786 / 1.792, 2.989 / 2.999 s; with MTP 1.276, 2.384, 3.433 s; prompt 468 (2B) or 464 vs 541 tokens;
%      decode 21.1 / 42.5 / 62.3 vs 32.1 / 65.2 / 107.2 ms per token. Screenshot steps (zhlocal.py, no
%      log-probabilities): exp_screen.sh shots-sizes, 21:43-22:49 (right: 2B 8/21 Tandem, 7/21 llama.cpp; 4B 16/21 and
%      9B 15/21 under both).

For Qwen3.5, the text-step latency reduction grows from 27\% at 2B to 34\% at 9B (Table~\ref{tab:sizes}). Checkpoint placement removes 14\% of prompt work at each size, while decode latency falls by 34\% and 42\%, respectively, as weight traffic increases. Screenshot gains are smaller because both engines run the same encoder; the 9B model's larger encoder leaves both at 18--20~s per screenshot step.

% AndroidControl moved to tab:workloads and the larger-workloads paragraph (\secref{sec:eval-e2e}).
% AndroidControl uses English tasks and places the rules in a separate system message. This gives llama.cpp a checkpoint at the user-message boundary. \sys serves the 300 test steps in 1.13~s on average, compared with 1.57~s for llama.cpp, with identical answers. Execution improvements and finer-grained prefix reuse within the user message provide the remaining speedup.
% src: androidcontrol/results/sc-ac-*.jsonl (exp_screen.sh ac, 2026-10-02 19:50-20:22, rt25 with all mechanisms, screen
%      verifiably on, order tan up0 up0 tan: 1.119 / 1.565 / 1.574 / 1.132 s, 28.3% less; identical answers in all four
%      runs (300/300); type accuracy 79.7%, step success 66.7% in all four runs; 366 vs 393 prompt tokens per step)

% \subsection{Other hybrid designs}\label{sec:eval-arch}
\subsubsection{Other hybrid designs}\label{sec:eval-arch}

\begin{table}[t]
\centering
\caption{LFM2-2.6B and Granite-4.0-H-Micro on the NPU, with the cooler: new prompt tokens and server time per text step of the replay trace, and power-collapse time during a 60-second idle window. Neither model has a draft head or a Gated DeltaNet layer. Granite's Mamba2 scan runs on the CPU in both engines.}
\label{tab:arch}
\footnotesize
\begin{tabular}{lcc}
\toprule
 & LFM2 & Granite \\
\midrule
Recurrent + attention layers & 22 conv.\ + 8 & 36 Mamba2 + 4 \\
\midrule
\multicolumn{3}{l}{\emph{New prompt tokens per step}} \\
\quad stock checkpoints & 581 & 610 \\
\quad \sys checkpoint policy & 518 & 534 \\
\multicolumn{3}{l}{\emph{Server time per step (s)}} \\
\quad llama.cpp & 0.95 & -- \\
\quad \sys, stock checkpoints & 0.88 & 2.48 \\
\quad \sys & 0.80 & 2.19 \\
\multicolumn{3}{l}{\emph{NPU power-collapsed (s of 60)}} \\
\quad votes held & 0 & 0 \\
\quad idle power release & 60 & 60 \\
\bottomrule
\end{tabular}
\end{table}
% src: research/route1/results/screen.log (exp_screen.sh arch, final build rt25, 2026-10-02 22:54-23:16, without
%      log-probabilities, screen verifiably on; "stock checkpoints" = no checkpoint rule, "\sys" = divergence-point
%      checkpoints and batch consolidation; LFM2 llama.cpp 0.939 / 0.962, stock 0.881 / 0.885, Tandem 0.797 / 0.802 s,
%      prompt_n 580.7 / 517.7; Granite stock 2.472 / 2.492, Tandem 2.188 / 2.188 s, prompt_n 609.7 / 533.6); arch.log
%      ar-lfm2-idle3000 / idle0 and ar5-granite (cDSP power-collapse time over 60 idle s: 61 / 60 and 0 s); LFM2_DET.md
%      for the fix that made the LFM2 runs repeatable. From unrounded times: LFM2 policy 9.4%, against llama.cpp 15.9%;
%      Granite 11.9%.

Checkpoint placement also reduces prompt work for other recurrent layers (Table~\ref{tab:arch}). We test LFM2-2.6B~\cite{liquid2025lfm2}, which uses short convolutions, and Granite-4.0-H-Micro~\cite{ibm2025granite4}, which uses Mamba2~\cite{dao2024mamba2}; both retain some attention layers. The checkpoint policy reduces new prompt tokens by 11--12\%, lowering step latency by 9\% for LFM2 and 12\% for Granite. With the kernel changes included, LFM2 has 16\% lower step latency than llama.cpp. Granite's scan remains on the CPU, so its rows compare checkpoint policies within \sys. On both models, releasing idle votes allows power collapse throughout a 60-second idle window.

% \subsection{Second device}\label{sec:eval-device}
\subsubsection{Second device}\label{sec:eval-device}

On the Snapdragon 8 Elite Gen 5, \sys serves text steps in 1.21~s versus llama.cpp's 1.66~s, a 27\% latency reduction. Screenshot steps fall to 3.82~s versus 4.54~s, a 16\% reduction. The kernels require only recompilation for v81, and both servers run without root access. Cross-device answers are identical for llama.cpp and differ in two free-text summaries for \sys; action outputs are unchanged.
% src: research/route1/results/dev2/dev2_tail.log (dev2/exp_dev2_tail.sh, OnePlus 15 CPH2749 / SM8850 / Hexagon v81,
%      OS 16.0.9.400, 2026-10-02 10:30-11:51, passively cooled with every run starting below 42 C, screen on, without
%      log-probabilities; rt25 tree built for v81 with all mechanisms and upstream b11200 + work-queue fix for v81):
%      text steps Tandem 1.195 / 1.215 / 1.225 / 1.215 s (mean 1.212), llama.cpp NPU without MTP 1.677 / 1.676 / 1.641
%      / 1.628 s (1.655) in alternating runs (run U7 aborted with dspqueue_read failed 0x2e and was replaced by U12 at
%      11:53, dev2/exp_dev2_relcheck.sh); llama.cpp with MTP 2.132 s, GPU 5.178 s, CPU 8 threads 11.796 s; app set with
%      screenshots Tandem 3.82 / 3.81 s, llama.cpp 4.55 / 4.53 s, 20/21 correct under both; answers against the first
%      phone: Tandem 59/61 (two done summaries, same actions), llama.cpp 61/61, prompt tokens 61/61

\subsubsection{Foreground apps}\label{sec:eval-interference}

\begin{table*}[t]
\centering
% \begin{updblock}
\caption{Server time per step (s) on the replay trace while another app runs in the foreground, and the slowdown relative to the home screen; each scenario was measured in one session with alternating runs.}
\label{tab:interference}
\small
\begin{tabular}{lcccc}
\toprule
Foreground app & llama.cpp & \sys & Gain & Slowdown (llama.cpp / \sys) \\
\midrule
Home screen & 1.99 & 1.44 & 28\% & -- \\
Video playback & 2.03 & 1.44 & 29\% & 1.9\% / 0.4\% \\
Feed scrolling & 2.09 & 1.45 & 31\% & 5.3\% / 1.1\% \\
Camera preview & 2.41 & 1.68 & 30\% & 21\% / 17\% \\
\bottomrule
\end{tabular}
% \end{updblock}
\end{table*}
% src: research/route1/results/gaps/paper_numbers.txt (gaps/exp_interf.sh, 2026-10-18 01:56-02:51, final build with all
%      mechanisms, cooler, screen verifiably on, without log-probabilities, T U U T per scenario): home 1.437 vs 1.988 s,
%      video 1.443 vs 2.026 s (a 200 s 1080p30 H.264 test video in the stock player, playback checked by screenshots),
%      scrolling 1.453 vs 2.094 s (the Zhihu feed swiped every 1.5 s), camera 1.679 vs 2.405 s (stock camera preview);
%      answers identical in every run (61/61). Feed frames (dumpsys gfxinfo): 5 of 8,187 janky without a server, 3 of
%      8,227 and 6 of 8,174 with Tandem, 1 of 11,641 and 8 of 11,763 with llama.cpp.

% \begin{updblock}
An agent's model shares the phone with the app it operates and with other foreground work. Table~\ref{tab:interference} runs the replay trace while another app holds the screen. Video playback and feed scrolling slow \sys by at most 1\% and llama.cpp by up to 5\%. The camera preview, which moves image data through memory continuously, slows decoding in both runtimes and adds 17\% to \sys's step and 21\% to llama.cpp's. \sys's reduction stays between 28\% and 31\% in all four cases, with identical answers. In the other direction, model execution does not disturb the foreground app: while the feed scrolls, 0.01--0.07\% of its frames miss their deadline, compared with 0.06\% without a server.
% \end{updblock}

\subsection{Fidelity and overhead}\label{sec:eval-fidelity}

With the vision encoder fixed on the NPU, each tested serving or kernel optimization preserves action outputs on the 61-request text trace and the 21-screen app set when added or removed individually. Changes in batching or speculative execution reword at most one free-text summary per trace. For text requests, the fused operator, data-movement paths, verification kernel, trimmed draft head and idle power release also preserve the measured first-token top-five log-probabilities.

Both \sys and the llama.cpp NPU baseline use the same NPU encoder. CPU and NPU encoder paths can produce different answers. The memory cost of \sys is 201\,MiB of rollback planes, 68.8\,MB for the draft head and one additional checkpoint of about 50\,MiB. The prompt cache is capped at 512\,MiB.
% src: research/route1/results/abl3 (the rt25 ablation, block 1, with log-probabilities; research/route1/paper_data/
%      fidelity.csv): added in turn, P, C, G, W, R, D and I keep 61/61 answers and first-token top-5 log-probs on the
%      replay trace and 21/21 on the screenshot steps, except R on the screenshot steps (21/21 answers, 0/21 top-5); M, H,
%      T and X each keep 60/61 answers (a done summary with the same action and success flag: ep00026/4 for M and H,
%      ep00017/10 for T and X) and 21/21 screenshot answers. Removed from the full configuration: -H, -T and -M each
%      reword the summary of one screenshot step (weather_main), -T, -M and -X one text summary (ep00017/10-11); no action
%      changes. -V (encoder on the CPU): 15/21 screenshot answers identical, 4 actions differ; 20/21 right with the NPU
%      encoder, 18/21 with the CPU encoder. Idle release: IDLE_POWER.md sec. 11 (61/61 answers and top-5 log-probs)

% Related work, Route 1 draft revision 2 (2026-09-30), STYLE_MOBISYS.md sec. 9: bold run-in categories, each closing with
% how Tandem differs. Replaces sections/related.tex (previous story).
\section{Related Work}\label{sec:related}

\noindent\textbf{LLM inference on mobile NPUs.} llm.npu accelerates prefill with chunk-sharing graphs and shadow execution of outliers~\cite{xu2025llmnpu}; HeteroLLM partitions computation across the NPU, GPU and CPU~\cite{chen2025heterollm}; PowerInfer-2 exploits activation sparsity~\cite{xue2024powerinfer2}; T-MAN uses table lookups for low-bit decode~\cite{wei2025tman}; and ShadowNPU offloads sparse attention~\cite{yin2026shadownpu}. These systems target full-attention models. Measurement studies examine mobile-NPU efficiency~\cite{cai2026npu} and test-time scaling~\cite{hao2026scaling}. Mobile deployments use engines such as MNN~\cite{jiang2020mnn}, MLC LLM~\cite{mlc2023mlcllm} and llama.cpp~\cite{ggml2023llamacpp,ggml2026hexagon}. \sys addresses recurrent-state management and the request pattern of a resident phone agent.

\noindent\textbf{Serving hybrid and recurrent models.} 
% Marconi checkpoints hybrid state at radix-tree branch points and admits entries according to expected reuse and compute savings~\cite{pan2025marconi}. 
\zm{Marconi checkpoints hybrid states at radix-tree branch points, uses estimated reuse for cache admission, and accounts for compute savings per byte in eviction~\cite{pan2025marconi}.}
Sparse prefix caching chooses checkpoints through dynamic programming over prefix-overlap distributions~\cite{shirokikh2026sparse}. GPU systems support speculative rollback through verification kernels and tree scans~\cite{wang2024mamballama,wu2025stree}. 
% \sys uses observed divergence in a single request stream to place checkpoints and separates chunked NPU prefill from the short tail that produces rollback states.
\zm{\sys uses cost-calibrated checkpoint spacing for a single agent's requests and preserves chunked NPU prefill by confining rollback-state generation to a short tail.}

\noindent\textbf{Speculative decoding.} Speculative decoding verifies the drafts of a small model in one pass of the large one~\cite{leviathan2023speculative}. MTP heads~\cite{gloeckle2024multitoken} and feature-level drafters~\cite{li2025eagle3} draft from the target model itself, and FR-Spec restricts drafting to frequent tokens~\cite{zhao2025frspec}, which \sys's trimmed draft head follows. 
% Agent-X observes that verifying two tokens on an edge device costs nearly as much as decoding them one by one~\cite{chung2026agentx}. On the phone NPU, \sys removes this cost with a kernel that reads each weight once for all drafts.
\zm{Agent-X~\cite{chung2026agentx} improves on-device agent inference through prompt rewriting for KV-cache reuse and example-based drafting. \sys instead targets recurrent-state reuse and reduces Hexagon verification overhead by sharing weight reads across drafts.}

\noindent\textbf{On-device GUI agents.} GUI agents automate app interactions using language models~\cite{wen2024autodroid,wang2024mobileagent,qin2025uitars,liu2024autoglm}. On-device approaches reduce model calls~\cite{wen2025autodroidv2}, shrink the policy~\cite{wang2026stepxedge,jiang2026openphone}, divide work between device and cloud~\cite{yi2026ecoagent}, learn through exploration~\cite{huang2026mobileexplorer} or reuse prefixes across screens~\cite{zhang2026vlmcache}. Other work reduces power during mobile inference~\cite{chu2026actbefore}. \sys accelerates individual calls of a hybrid policy and releases the NPU's power votes between request bursts.

% Discussion and conclusion, Route 1 draft revision 2 (2026-09-30), STYLE_MOBISYS.md sec. 9: run-ins that go condition,
% consequence, path; conclusion with the headline numbers.
\section{Discussion}\label{sec:discussion}

\noindent\textbf{Serving policies.} 
% A runtime can use the structure of an agent's requests to decide which completed work to retain. In our workload, fixed instructions and slowly changing history make the overlap between consecutive requests informative about later reuse. This guides both checkpoint placement and batch boundaries: saving a reusable state has value, while saving an unused one adds a copy and a model pass. 
% \zm{Checkpoint gains also depend on prompt format: AndroidControl's separate system message already allows llama.cpp to checkpoint at the user-message boundary, leaving less shared text to recover (\secref{sec:eval-e2e}).}
% Checkpoint placement, batch consolidation and idle power release can be adopted independently of \sys's kernels.
\zm{Checkpoint gains depend on prompt structure and may be smaller when the baseline already preserves reusable prefixes (\secref{sec:eval-e2e}). Checkpoint placement, batch consolidation and idle power release can be adopted independently of \sys’s kernels.}

\noindent\textbf{Other models and accelerators.} Divergence-point checkpoints require a server that can save and restore recurrent state. Idle power release requires session-level power control. Both mechanisms apply to the short-convolution and Mamba2 hybrids we evaluate. The speculative path depends more closely on the model and hardware: rollback-aware prefill uses Gated DeltaNet's chunked recurrence, and the MTP head drafts from the target's hidden states. The Hexagon kernels carry from v79 to v81 through recompilation; another accelerator would need corresponding matrix and vector implementations.

\noindent\textbf{Screenshot cost.} 
% Image encoding and image-token prefill dominate screenshot requests, limiting the benefit of text reuse and faster decode. The image-budget study quantifies the latency--accuracy tradeoff (\secref{sec:eval-image}). Screens requiring fine visual detail may favor a larger budget. Selecting a budget from screen content and accelerating the encoder offer further reductions in this part of the step.
\zm{Screens requiring fine visual detail may favor a larger image-token budget, despite the latency savings from reducing it (\secref{sec:eval-image}). Adapting the budget to screen content and accelerating the vision encoder could further reduce screenshot latency.}

\section{Conclusion}\label{sec:conclusion}

Hybrid models reduce the memory needed to run GUI agents on a phone. \sys adapts their serving to changing prompts, short action outputs and idle intervals: it preserves reusable recurrent states across steps, supports speculative execution while retaining fast prompt processing, and releases idle NPU resources without unloading the model. On a Snapdragon 8 Elite, it reduces text-step latency by 30\%, screenshot-step latency by 16\% and energy per text step by 33\% relative to llama.cpp's fastest NPU configuration. Standby power with the model loaded falls from 0.76 to 0.22~W. The text-step latency reduction remains 27\% on the next NPU generation.

\label{lastbodypage}
\bibliographystyle{mlsys2025}
\bibliography{references}

\end{document}